\documentclass{article}

\usepackage{microtype}
\usepackage{graphicx}
\usepackage{booktabs} 

\usepackage{hyperref}

\usepackage[accepted]{icml2023}

\usepackage{amsmath}
\usepackage{amssymb}
\usepackage{mathtools}
\usepackage{amsthm}

\usepackage[capitalize,noabbrev]{cleveref}

\theoremstyle{plain}

\theoremstyle{definition}

\theoremstyle{remark}

\usepackage[textsize=tiny]{todonotes}
\usepackage{stmaryrd,enumitem}
\usepackage{xspace}
\usepackage{soul}

\usepackage{multirow}
\usepackage{array}
\usepackage{makecell}
\usepackage{comment}
\usepackage[position=bottom]{subfig}
\newtheorem{defn}{\textbf{Definition}}
\newtheorem{prob}{\textbf{Problem}}

\newtheorem{thm}{\textbf{Theorem}}

\newtheorem{props}{\textbf{Proposition}}
\newtheorem{cor}{\textbf{Corollary}}
 \newtheorem{lem}{\textbf{Lemma}}
\setlist{nolistsep,leftmargin=*}

\newcommand{\namemodelnew}{\textsc{Grafenne}\xspace}
\newcommand{\FP}{\textsc{FP}\xspace}

\newcommand{\gnn}{\textsc{Gnn}\xspace}

\newcommand{\gnns}{\textsc{Gnn}s\xspace}
\newcommand{\graphsage}{\textsc{GraphSage}\xspace}
\newcommand{\gcnmf}{\textsc{GcnMf}\xspace}
\newcommand{\pagnn}{\textsc{PaGNN}\xspace}
\newcommand{\gcn}{\textsc{Gcn}\xspace}
\newcommand{\gat}{\textsc{Gat}\xspace}
\newcommand{\gin}{\textsc{Gin}\xspace}
\newcommand{\fp}{\textsc{Fp}\xspace}
\newcommand{\nm}{\textsc{Nm}\xspace}

\newcommand{\fate}{\textsc{Fate}\xspace}

\newcommand{\m}{\mathbf{m}\xspace}
\newcommand{\M}{\mathbf{M}\xspace}

\newcommand{\h}{\mathbf{h}\xspace}
\newcommand{\W}{\mathbf{W}\xspace}
\newcommand{\w}{\mathbf{w}\xspace}

\newcommand{\X}{\mathbf{X}\xspace}

\newcommand{\x}{\mathbf{x}\xspace}

\newcommand{\CG}{\mathcal{G}\xspace}

\newcommand{\CV}{\mathcal{V}\xspace}
\newcommand{\CE}{\mathcal{E}\xspace}

\newcommand{\cx}{\mathbf{x}\xspace}

\newcommand{\ch}{\mathbf{h}\xspace}

\newcommand{\alt}{alt}

\icmltitlerunning{\hfill \namemodelnew: Learning on Graphs with Heterogeneous and Dynamic Feature Sets \hfill \thepage}

\begin{document}

\twocolumn[
\icmltitle{\namemodelnew:  Learning on Graphs with Heterogeneous and Dynamic Feature Sets}



\icmlsetsymbol{equal}{*}

\begin{icmlauthorlist}
\icmlauthor{Shubham Gupta}{equal,iitd}
\icmlauthor{Sahil Manchanda}{equal,iitd}
\icmlauthor{Sayan Ranu}{iitd}
\icmlauthor{Srikanta Bedathur}{iitd}
\end{icmlauthorlist}

\icmlaffiliation{iitd}{Department of Computer Science and Engineering, IIT Delhi}

\icmlcorrespondingauthor{Shubham Gupta}{shubham.gupta@cse.iitd.ac.in}
\icmlcorrespondingauthor{Sahil Manchanda}{sahil.manchanda@cse.iitd.ac.in}

\icmlkeywords{Graph Machine learning, Graph Neural Networks, Lifelong Learning, Continual Learning, Dynamic features, Diverse features, Missing features, Data sparsity, Streaming Graph, Expressivity, Inductive GNNs}

\vskip 0.3in
]



\printAffiliationsAndNotice{\icmlEqualContribution} 

\begin{abstract}
 Graph neural networks (\gnns), in general, are built on the assumption of a static set of features characterizing each node in a graph. This assumption is often violated in practice. 
Existing methods partly address this issue through \textit{feature imputation}. However, these techniques \textit{(i)} assume uniformity of feature set across nodes, \textit{(ii)} are transductive by nature, and \textit{(iii)} fail to work when features are added or removed over time. In this work, we address these limitations through a novel \gnn framework called \namemodelnew. \namemodelnew performs a novel \textit{allotropic} transformation on the original graph, wherein the nodes and features are decoupled through a bipartite encoding. Through a carefully chosen message passing framework on the allotropic transformation, we make the model parameter size independent of the number of features and thereby inductive to \textit{both} unseen nodes and features. We prove that \namemodelnew is at least as expressive as any of the existing message-passing \gnns in terms of Weisfeiler-Leman tests, and therefore, the additional inductivity to unseen features does not come at the cost of expressivity. In addition, as demonstrated over four real-world graphs, \namemodelnew empowers the underlying \gnn with high empirical efficacy and the ability to learn in continual fashion over streaming feature sets.

\end{abstract}
\section{Introduction and Related Work}

Graph Neural Networks (\gnns) have witnessed immense popularity in modeling topological data. \gnns have produced state-of-the-art results in molecular property prediction~\cite{graphormer,graphgps}, protein function prediction~\cite{graphsage,graphreach,pgnn}, modeling of physical systems~\cite{lgnn,gnode,benchmarking,rigidbody}, traffic forecasting~\cite{frigate,neuromlr}, material discovery~\cite{stridernet},  learning combinatorial algorithms~\cite{gcomb,greed,wsdmim,ilp}, and graph generative modeling~\cite{graphgen,graphrnn,tigger,digress}. The effectiveness of \gnns is closely associated with the availability of high-quality input node features~\cite{fp}. An inherent assumption in existing \gnns is the availability of \textit{all} features for each node in the graph. In practice, this assumption is often violated producing \textit{heterogeneous} and \textit{dynamic} feature sets. To motivate, we list two commonly occurring scenarios.
\looseness=-1

\noindent
$\bullet$ \textbf{Applicability of features:} In graphs with heterogeneous features, the feature set characterizing node $v$ may not be relevant for node $u$. As an example, consider a co-purchase graph over items in an e-commerce database. While the feature \textsc{Cpu} clock-speed is relevant for smartphones, it does not apply to smartphone covers. While one may homogenize the features set across all nodes by attributing a special value to denote non-relevant features, it significantly enlarges feature dimensionality leading to inefficiencies in the modeling, computational and storage components.\\
\noindent
$\bullet$ \textbf{Feature set refinement:} Feature sets may get altered over time~\cite{snapnets, 10.1145/3209978.3210129}. Consider the evolution of smartphones into the foldable form factor. While a feature characterizing the form-factor is required in today's context, it was not conceivable five years back. Similarly, in a dating network, users might choose to remove/hide personal attributes related to income, education, and occupation at a later stage after initial sign-up. Furthermore, the dating app might ask for covid vaccination status in the post-pandemic era. Since the number of model parameters in \gnns is a function of the input node feature dimension, when the feature set changes, the entire model needs to be retrained from scratch. The ideal solution lies in decoupling the \gnns parameters with feature set size and \textit{continually} \textit{adapting} the existing model to new features without forgetting the past. 
\looseness=-1


\subsection{Existing Works}
The closest work to ours is topology-aware feature imputation over missing features~\cite{gcnmf,pagnn,fp}\footnote{{Topology-unaware methods~\cite{wu2021towards} are not well-suited for the task due to their inability to model node-dependencies (See. Appendix Sec.~\ref{app:comp_FATE})}}. In feature imputation, the value of a non-existing feature is predicted based on other existing features and the graph topology. Feature imputation, however, is not adequate for the proposed problem.

\begin{itemize}
\item {\textbf{Feature relevance:}} Feature imputation assumes that a feature is relevant, but missing. Heterogeneous feature sets surfaces a different problem where a feature itself is not relevant and hence imputation is an irrational task. 

\item {\textbf{Transductive modeling:}} Topology-aware feature imputation algorithms, such as \gcnmf \cite{gcnmf}, \pagnn \cite{pagnn} and \fp \cite{fp}, require computation of the Graph Laplacian, rendering them incapable of modeling unseen nodes, i.e., nodes not observed in training. 

\item {\textbf{Lacking ability for continual learning:}} As discussed above, it is natural for datasets to update feature sets to stay in sync with their evolution. This necessitates developing a \textit{continual learning} framework for a streaming node feature scenario that avoids \textit{catastrophic forgetting} on the portion of the graph that is unaffected in the update. Existing algorithms for topology--aware feature imputation do not support this need due to being transductive. On the other hand, algorithms for continual learning on graphs \cite{wang2022lifelong,continualgnn,liu2021overcoming} perform continual learning only over nodes with homogeneous feature sets.
\looseness=-1
\end{itemize}

    
    

\vspace{-0.05in}
\subsection{Contributions}
\vspace{-0.05in}
In this work, we propose \underline{GRA}ph  \underline{FE}ature  \underline{N}eural  \underline{NE}twork (\namemodelnew) to address the above-highlighted gaps. \namemodelnew is built on the following novel contributions:
\vspace{-0.05in}
\begin{itemize}
    \item \textbf{Continual, assumption-free and \gnn-agnostic modeling:} \namemodelnew transforms the input graph into a graph consisting of two disjoint sets of \textit{graph nodes} and \textit{feature nodes}. Through a novel message passing scheme across these nodes, \namemodelnew ensures three key properties. First, the number of model parameters is independent of the number of nodes or features. Hence, it is inductive to both unseen nodes and features. This enables an easy transition to \textit{continual} learning. Second, the flexible transformation and message passing scheme can mimic any of the existing \gnn architectures. Third, it bypasses the need to impute features and thereby imbibing the ethos of heterogeneous feature sets.

    \item \textbf{Expressivity:} We prove that given any \gnn of $k$-WL expressivity, they can be embedded into the \namemodelnew framework to retain their full expressive power under $k$-WL while also imbibing the above mentioned properties.
    \looseness=-1
    
    
    
    \item \looseness=-1 \textbf{Empirical evaluation:} Extensive experiments on diverse real-world datasets 
 establish that \namemodelnew consistently outperforms baseline methods across various levels of feature scarcity on both homophilic as well as heterophilic graphs. Further, the method is robust to extreme low availability of node features.

    
\end{itemize}
\vspace{-0.10in}

\section{Preliminaries}
\begin{defn}[Graph]
\label{def:input_graph}
\textit{A graph is defined as $\CG=(\CV,\CE,\X)$ over node and edge sets $\CV$ and $\CE=\{(u,v) \mid u,v \in \mathcal{\CV}\}$  respectively. $\X \in \mathbb{R}^{\mid V\mid \times \mid F\mid}$ is a node feature matrix where $F$ is the set of all features in graph $\CG$. $F$ is assumed to be available for all nodes $v \in \CV$.}
\end{defn}
Assuming $\x_v \in \mathbb{R}^{\mid F\mid}$ as input feature vector for every node $v \in \CV$, the $0^{th}$ layer embedding of node $v$ is denoted as:

\begin{equation}
\h_v^0 = \x_v \;\; \forall v \in \CV
\label{eq:gnn_initialize}
\end{equation}

To compute the $\ell^{th}$ layer representation of node $v$, \gnns compute the message from its neighbourhood $\mathcal{N}_v = \{u \mid (u,v) \in \CE\}$ and aggregate them as follows:
\begin{equation}
\label{eq:gnn_msg}
\m_v^\ell(u) = \text{\textsc{Msg}}^\ell(\h_u^{\ell-1},\h_v^{\ell-1}) \;\forall u \; \in \mathcal{N}_v
\end{equation}

\begin{equation}
\label{eq:gnn_msg_agg}
\overline\m_v^\ell = \text{\textsc{Aggregate}}^l(\{\!\!\{\m_v^l(u), \forall u \in \mathcal{N}_v\}\!\!\})   
\end{equation}

where $\text{\textsc{Msg}}^l$ and $\text{\textsc{Aggregate}}^l$ are either pre-defined functions (Ex: \textsc{MeanPool}) or neural networks (\gat~\cite{gat}). $\{\!\!\{\ldots\}\!\!\}$ denotes a multi-set; it is a multi-set since the same message may be received from multiple neighbors.
 Finally, \gnns compute the $l^{th}$ layer representation of node $v$ as follows:

\begin{equation}
    \label{eq:gnn_combine}
    \h_v^\ell = \text{\textsc{Combine}}^l(\h_v^{\ell-1},\overline\m_v^\ell)
\end{equation}

where $\text{\textsc{Combine}}$ is a neural network. The node representations of the final layer, denoted as $\ch_v$, are used for downstream tasks such as node classification, link prediction, etc.
\looseness=-1

\textbf{Assumptions:} As summarized in the above generic framework, all node representations undergo the same transformation in each layer. While this design allows \gnns to decouple the number of model parameters from the node set size $|\CV|$, it forces the parameter size to be a function of the node representation dimension. 
Moreover, node features $\x_v$ are tightly coupled with node representation $\h_v^\ell$ in each hidden layer $\ell$ (See  Eq.~\ref{eq:gnn_initialize} and Eq.~\ref{eq:gnn_combine}). Consequently, if the feature set size is heterogeneous, or dynamic due to changes over time, \gnns are either inapplicable, or requires re-training from scratch following each feature set update. Our objective is to remove this assumption and mitigate the resultant shortcomings.

\section{Problem Formulation}
First, we redefine graph (Def.~\ref{def:input_graph}) by relaxing the constraint that all nodes should consist of same set of input features:

\begin{defn}[Graph with heterogeneous feature set]
\label{def:graph}
\textit{A graph 
is defined as  $\CG=(\CV,\CE,\X)$ where $\CV$ is a set of nodes and $\CE=\{(u,v) \mid u,v \in \mathcal{V}\}$ is a set of edges. $\X = \left[\x_1,\x_2 \ldots \x_{\mid \CV\mid}\right]$ is a collection of node feature vectors where $\x_v \in \mathbb{R}^{\mid F_v \mid}, \forall v \in \CV $. $F_v$ is the set of features available at node $v$ and $F= \bigcup _{v \in \CV}F_v$ is the union of available features across all nodes in the graph. }
\end{defn}

\noindent
We note that $\X$ is not a matrix since the dimension of each $\x_v\; \forall v \in \CV$ can be different. Moreover, each dimension in $\x_v$ may have a different meaning for every $v$. Thus, existing GNNs cannot train on these graphs  without including missing-value imputation as an intermediate step. Motivated by this, we now define the following novel formulation.

 
\begin{prob}[\gnn for graphs with heterogeneous features]\hfill
\label{prob:1}

\noindent
\textbf{Input:} \textit{Given a graph $\CG$ (Def.~\ref{def:graph}), let $Y:\mathcal{V} \rightarrow \mathbb{R}$ be a hidden function that maps a node to a real number\footnote{It is easy to adapt $Y(v)$ for edge or graph level tasks by learning an aggregation function over its constituent node representations.}. $Y(v)$ is known to us only for subset $\CV_l \subset \CV$ and may model some downstream task such as node classification, or link prediction.
\looseness=-1}


\noindent
\textbf{Goal:} \textit{Learn parameters $\Theta$ of a graph neural network, denoted as $\gnn_{\Theta}$, that predicts $Y(v)$, $\forall v\in \CV_l$ accurately. We focus on inductive learning so that $\gnn_{\Theta}$ can predict on unseen graphs (and nodes).}


\end{prob}

Prob.~\ref{prob:1} is formulated in the context of static graphs with heterogeneous feature sets. 
  It does not consist any temporal characterisation. 
  We further generalize Def.~\ref{def:graph} to allow temporal characteristics.
\begin{defn}[Dynamic graphs]
\label{def:streaming_graph}
\textit{A dynamic (or streaming) graph is a sequence of graph (as per Def.~\ref{def:graph}) snapshots recorded at consecutive timestamps $t=1,2 \ldots$ and represented as $\overrightarrow{\mathcal{G}} = (\CG_1,\CG_2 \ldots)$ where $\CG_t = (\CV_t,E_t,\X_t)$. 
}
\end{defn}
In Def.~\ref{def:streaming_graph},  nodes/edges/features can be added or existing ones can be deleted in consecutive snapshots, i.e., $\CV_{t+1} = \CV_t + \Delta \CV_t,$ $E_{t+1} = E_t + \Delta E_t$ and $\X_{t+1} = \X_t + \Delta \X_t$. We note that different subsets of  features can be added/removed for all nodes or a subset of nodes during each update. To ease the notational burden, we overload the notation of $\Delta\CV_T$ to denote all nodes that were either deleted, added, or underwent a feature change. Each of the three update types can be distinguished with an appropriate indicator variable.


Following Def.~\ref{def:streaming_graph}, we write $\CG_t=\CG_{t-1}+\Delta \CG_t$ where $\Delta \CG_t = (\Delta \CV_t, \Delta E_t, \Delta X_t)$. The sequential graph updates $\Delta \CG_t$ may contain new patterns, thus making $\gnn_{\Theta_1}$ trained using $\CG_1$,  ineffective over $\CG_2$, $\CG_3$, $\ldots, \CG_t$. However, re-training $\gnn_{\Theta_t}$ from scratch on $\CG_t$ is computationally expensive as well. A vanilla solution of online learning resides in updating $\Theta_{t-1}$ only on $\Delta \CG_t$. This, however, may lead to \textit{catastrophic forgetting}~\cite{parisi2019continual} since the learned parameters will be biased towards new patterns and drift away from patterns learned earlier.
This motivates us to develop a \textit{continual training} framework where learned parameters perform effectively on $\Delta \CG_t$ while retaining information about $\CG_t - \Delta \CG_t$ as well. This leads us to define the following problem statement.
\begin{prob}[Continual Learning over \gnns]
\label{prob:2}
\textit{We extend Prob.~\ref{prob:1} for dynamic graphs with the following   objectives:}
\vspace{-0.05in}
\begin{itemize}
\item \textit{{\bf Training accuracy:} Given $\overrightarrow{\CG}=\{\CG_1,\ldots,\CG_t\}$, learn $\{\Theta_1,\ldots,\Theta_t\}$ such that $\forall v\in \CV_{t_l},\;\gnn_{\Theta_t}$ is accurate; $\CV_{t_l}\subseteq\CV_t$ is the set of training nodes. }

\item\textit{{\bf Computational efficiency:} Learning $\Theta_t$ should be significantly more efficient than retraining from scratch on $\CV_{t_l}$.}
\looseness=-1
\end{itemize}
We achieve the above objectives by grounding our update mechanism for $\Theta_t$ predominantly on $\Delta\CV_t\cap\CV_{t_l}$. The next section details our methodology.
\end{prob}





\section{Proposed \gnn framework: \namemodelnew}
\namemodelnew constitutes of three major components: \textbf{(1)} An \textit{allotropic} graph transformation that enables decoupling of model parameter-size from the number of features, \textbf{(2)} a novel generic message-passing mechanism on transformed graph that is provably as expressive as performing message passing on the original graph, and \textbf{(3)} a continual learning framework to adapt to new nodes, edges and features efficiently and without catastrophic forgetting.
\looseness=-1
\begin{figure}[t]
\centering
\includegraphics[scale=0.27]{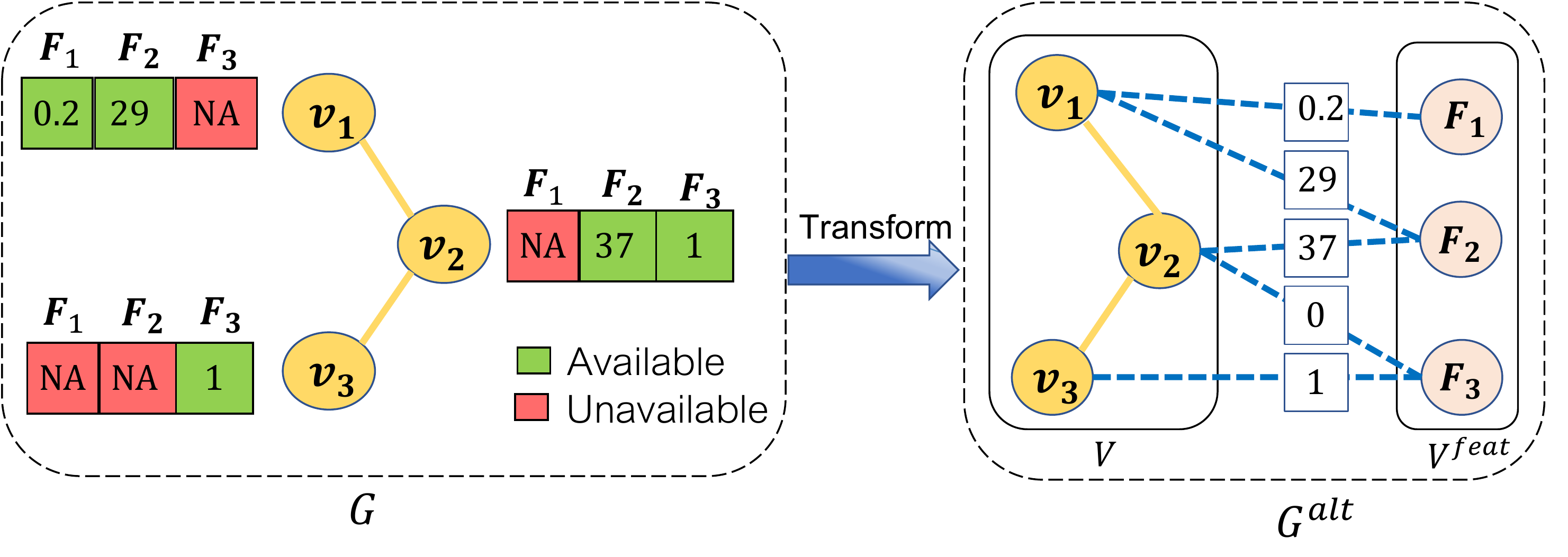}
\vspace{-0.1in}
\caption{Transformation of input graph $\CG$ to its allotropic form $\CG^{\alt}$. In $\CG$, the features marked in green are available for the corresponding node.}
\label{fig:transformation}
\end{figure}

\subsection{Graph Transformation}
Given graph snapshot (Def. \ref{def:graph}) $\CG=\left(\CV,\CE,\X\right)$, we construct its \textit{allotropic} version $\CG^{\alt}= \left(\CV^{\alt},\CE^{\alt}\right)$. $\CV^{alt}=\CV \cup \CV^{feat}$, where in addition to the original nodes $\CV$, we add a node for each unique feature in $\CG$. Formally, 
\begin{equation}
\nonumber
\CV^{feat}=F=\left\{  {f}\mid f \in \cup _{\forall v \in \CV} F_v \right\}\;\;\hfill \text{(Recall Def.~\ref{def:graph})}
\end{equation}
We call $\CV^{feat}$ the \textit{feature nodes}. A regular node $v\in\CV$ is connected to a feature node $f\in \CV^{feat}$ if $f$ characterizes $v$, i.e., $f\in f_v$. The weight of this edge is the value of feature $f$ in node $v$. In addition, we also retain all the edges among the original nodes. Thus $\CE^{\alt} = \CE \cup \CE^{feat}$, where:
\begin{equation}
\nonumber
   \CE^{feat} = \left\{\left(v,f,\x_{v}[f]\right) \mid v \in \CV, f \in F_v\right\}
\end{equation}
Here, $\x_{v}[f]{\in\mathbb{R}}$ refers to the edge weight/feature value between original graph node $v$ and  feature $f \in \CV^{feat}$. 
 Fig.~\ref{fig:transformation} illustrates the transformation process.

In $\CG^{alt}$, the feature nodes only have graph nodes as neighbors. In contrast, graph nodes have both feature nodes as well as other graph nodes in their neighborhood. To distinguish between these two types, we define the notions of \textit{graph neighborhood} and \textit{feature neighborhood}. 
\begin{defn}[Feature Neighborhood]
\textit{The feature neighborhood of a node $v\in \CV^{alt}$ is defined as $\mathcal{N}_v^{feat} = \left\{u\mid \left(u,v\right) \in {\CE ^{feat}},\;v\in\CV^{feat}\right\}$.}
\end{defn}
\begin{defn}[Graph Neighborhood]
\textit{For a given node $v \in \CV^{\alt}$, the graph neighbourhood $\mathcal{N}_v^{\CG}=\left\{u \;{\mid}\left(u,v\right) \in {\CE},\;v\in\CV\right\}$ consists of only graph nodes.}
\end{defn}

Note that $\forall v\in\CV^{feat},\:\mathcal{N}_v^{\CG}=\emptyset$.
\subsection{Message Passing Layer for $\CG^{\alt}$}
\label{sec:msg_passing}
Our goal is to perform message passing on $\CG^{\alt}$ in order to learn rich representations for graph nodes $v\in \CV$ such that: \textbf{(1)} Attribute information is captured from the feature nodes,  \textbf{(2)} Topological information is captured from the neighborhood defined over $\CE$, \textbf{(3)} It decouples the size of model parameters from the number of features, and \textbf{(4)} It theoretically guarantees that message passing on the allotropic form does not lead to reduction in expressive power when compared to executing a \gnn on $\CG$. 

To achieve our goals, the message passing scheme is broken down into three phases. A single layer of message passing is completed on the completion of these three phases.
\looseness=-1\\
\noindent
$\bullet$\textbf{Phase 1 -- Aggregating feature information:} Messages are sent from feature nodes $u\in\CV^{feat}$ to graph nodes $v\in\CV$ to aggregate edge weights (carrying the feature value). Formally,
\begin{alignat}{3}
\label{eq:fognn_feat_msg}
\m_v^{\ell}\left(u\right) &= \text{\textsc{Msg}}^{\ell}_{feat}\left(\h_u^{\ell-1},\h_v^{\ell-1}, e_{uv}\right)\\
\label{eq:fognn_feat_msg_agg}
\small{\overline\m_v^{\ell}} &= \small{\text{\textsc{Aggregate}}^{\ell}_{feat}\left(\left\{\!\!\left\{\m_v^{\ell}\left(u\right) \mid  u \in \mathcal{N}^{feat}_v\right\}\!\!\right\} \right) }  \\
    \label{eq:fognn_feat_combine}
    \h_v^{\ell} &= \text{\textsc{Combine}}_{feat}^{\ell}\left(\h_v^{\ell-1},\overline\m_v^{\ell}\right)
\end{alignat}

Here, $e_{uv}=\cx_v[u]$ is edge weight between feature node $u\in \CV^{feat}$ and graph node $v\in \CV$ in $\CG^{\alt}$. $\ell$ denotes the message-passing layer.

\noindent
$\bullet$ \textbf{Phase 2 -- Aggregating topological information:} Utilizing the information aggregated in the previous phase, exchange messages between graph nodes $u,v\in\CV$ and aggregate them as follows: 
\begin{alignat}{3}
\label{eq:fognn_gnode_gnode_msg}
\m_v^{\ell}\left(u\right) &= \text{\textsc{Msg}}^{\ell}_{\CG}\left(\h_u^{\ell},\h_v^{\ell}\right)\\
\label{eq:fognn_gnode_msg_agg}
\overline\m_v^{\ell} &= \text{\textsc{Aggregate}}^{\ell}_{\CG}\left(\left\{\!\!\left\{\m_v^{\ell}\left(u\right)\mid u \in \mathcal{N}^{\CG}_v\right\}\!\!\right\}\right)  \\
    \label{eq:fognn_gnode_combine}
    \h_v^{\ell} &= \text{\textsc{Combine}}_{\CG}^{\ell}\left(\h_v^{\ell},\overline\m_v^{\ell}\right)
\end{alignat}
 \noindent
$\bullet$ \textbf{Phase 3 -- Integrating attribute and topology information:} Messages are exchanged back from graph nodes $v\in\CV$ to feature nodes $u\in\CV^{feat}$ and aggregated as follows. 
\begin{alignat}{3}
\label{eq:fognn_gnode_feat_msg}
{\m_u^{\ell}\left(v\right)} &= {\text{\textsc{Msg}}^{\ell}_{\CG'}\left(\h_v^{\ell},\h_u^{\ell-1},e_{uv}\right)}\\
\label{eq:fognn_gnode_feat_msg_agg}
{\overline\m_u^{\ell}} &= \small{{\text{\textsc{Aggregate}}^{\ell}_{\CG'}\left(\left\{\!\!\left\{\m_u^{\ell}\left(v\right)\mid v \in \mathcal{N}^{feat}_u\right\}\!\!\right\}\right) }} \\
    \label{eq:fognn_gnode_feat_combine}
   { \h_u^{\ell}} &= {\text{\textsc{Combine}}_{\CG'}^{\ell}\left(\h_u^{\ell-1},\overline\m_u^{\ell}\right)}
\end{alignat}
 
\begin{figure*}[t]
\centering
\includegraphics[scale=0.30]{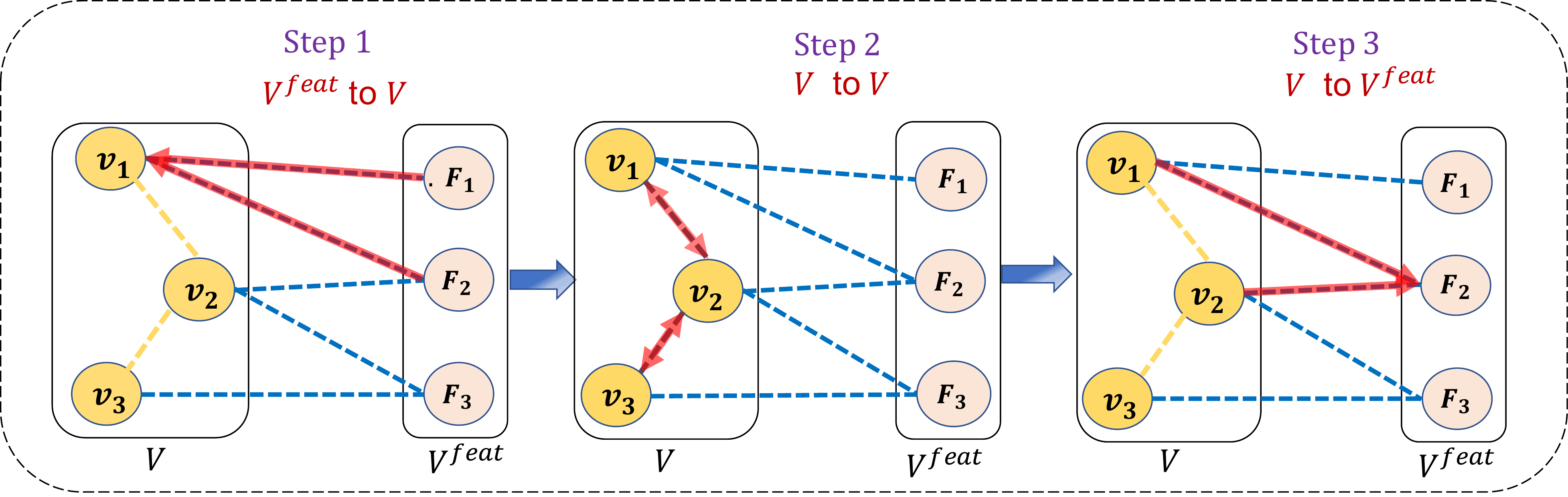}
\caption{A break-down of the message passing layer of \namemodelnew into its three phases. }
\label{fig:msg_passing}
\end{figure*}

 Fig.~\ref{fig:msg_passing} provides a visual depiction of the three phases of message passing in \namemodelnew. $\text{\textsc{Msg}}_{feat}^{\ell}$,  $\text{\textsc{Msg}}_{\CG}^{\ell}$,  $\text{\textsc{Msg}}_{\CG'}^{\ell}$,  $\text{\textsc{Aggregate}}_{feat}^{\ell}$,     $\text{\textsc{Aggregate}}_{\CG}^{\ell}$,  $\text{\textsc{Aggregate}}_{\CG'}^{\ell}$,  $\text{\textsc{Combine}}_{feat}^{\ell}$,  $\text{\textsc{Combine}}_{\CG}^{\ell}$ and $\text{\textsc{Combine}}_{\CG'}^{\ell}$ are neural network based functions with semantics defined in preliminaries section. Any existing \gnn can be used to implement these three phases since they are essentially exchanging information between nodes. This makes \namemodelnew a general and flexible method. We discuss our specific implementation in \S.~\ref{sec:specifics}.

  \noindent
\textbf{Initialization:} 
We use $\boldsymbol{0}$ as a feature vector for graph nodes $\CV$ in $\CG^{\alt}$ since they no longer have any node features, i.e., 
\begin{equation} 
    \h_v^0  = \boldsymbol{0} \; \forall \; v \in \CV
    \label{eq:zero_vector}
\end{equation}
For feature nodes $\CV^{feat}$, we set them either to a learnable vector initialized randomly or to a latent representation learnt in a pre-processing step.
Specifically, 
\begin{equation}
    \h_u^0 = \boldsymbol{w}_u \in \mathbb{R}^d\; \quad \forall \; u \in \CV^{feat}
    \label{eq:embedding}
\end{equation}

\subsubsection{Specifics}
\label{sec:specifics}
The neural networks in phase 1 and phase 3 are defined as attention-based aggregators. Since phase 2 involves message passing only among graph nodes, we can adopt the message passing scheme of any static \gnn \cite{graphsage,gat,gcn,higher_order_wl_gnn}. Some possible options are outlined in App.~\ref{app:phase2}. {Phases 1 and 3 may also be customized to different neural networks as per needs.}\\
\looseness=-1
\noindent \textbf{Phase 1: $\forall v \in \CV,\; u \in \mathcal{N}_v^{feat}$} 
\begin{alignat}{3}
\label{eq:phase1eq1}
    \m_v^{\ell}\left(u\right)&{=}\text{\textsc{LeakyReLU}}\left({\W_1^{\ell}}\h_v^{\ell-1}{ \bigparallel }{\W_2^{\ell}}\h_u^{\ell-1}{ \bigparallel  }{\w_3^{\ell}}e_{uv}\right)\\ 
    \alpha_{vu}&{ =} \frac{\exp\left({\w^{\ell}_4}^T\m^{\ell}_v\left(u\right)\right)}{\sum\limits_{u' \in \mathcal{N}_{v}^{feat}}\exp\left({\w^{\ell}_4}^T\m^{\ell}_v\left(u'\right)\right)}, \\ 
    \h_v^{\ell}&{ = } \text{\textsc{MLP}}\left( {\W^{\ell}_{5}}\h_v^{l-1}\bigparallel \sum_{u \in \mathcal{N}_{v}^{feat}}{ \alpha_{vu}{\W^{\ell}_{6}}\h_u^{\ell-1}} \right)
\end{alignat}
\looseness=-1
\noindent \textbf{Phase 3: $\forall u \in \CV^{feat},\; v {\in }\mathcal{N}_u^{\CG}$}
\begin{alignat}{3}
\nonumber
    {\m_u^{\ell}\left(v\right)}&{{=}\text{\textsc{LeakyReLU}}\left({\W_7^{\ell}}\h_u^{\ell-1}{ \bigparallel }{\W_8^{\ell}}\h_v^{l}{ \bigparallel }\w^{\ell}_9e_{uv}\right)}\\
    \nonumber
    \alpha_{vu}&{ =} {\frac{\exp\left({\w_{10}^{\ell}}^T\m^{\ell}_u\left(v\right)\right)}{\sum\limits_{v' \in \mathcal{N}_{u}^{\mathcal{G}}}\exp\left({\w^{\ell}_{10}}^T\m^{\ell}_u\left(v'\right)\right)}}\\
    \nonumber
   { \h_u^{\ell}}&{ = }\text{\textsc{MLP}}\left(\W^{\ell}_{11}\h_u^{\ell-1} \bigparallel {\sum_{v \in \mathcal{N}_{u}^{\CG}}}{ \alpha_{vu}\W^{\ell}_{12}\h_v^{l}} \right)
\end{alignat} 
\noindent
All weights matrices and vectors of the form $\W^{\ell}_i$ and $\w^{\ell}_i$ respectively are trainable parameters. $\bigparallel$ represents the \textit{concatenation} operator. Note that since each edge weight goes through an MLP, the proposed scheme is expressive enough to model scaling and translation factors.
\looseness=-1
\subsection{Theoretical Characterization}
With the formalization of our message-passing algorithm, we have a \gnn framework for graphs with heterogeneous feature sets. 
Next, we analyze its inductivity, expressivity, and complexity.
\looseness=-1
\subsubsection{Inductivity}
\begin{props}{
\textit{\namemodelnew is inductive to both unseen features and nodes, i.e., once trained, it is capable of producing representations for unseen nodes with unseen features.}}
\label{props:inductivity}
\end{props}
Existing \gnns are not feature-inductive as they must be re-trained if a new feature is added to the input graph. \namemodelnew decouples features from \gnns's parameters by treating them as nodes in $\CG^{alt}$. Moreover, \namemodelnew learns aggregation functions over feature nodes to compute the graph node representations in Phase-1. These aggregation functions are independent of the number of features available to the target node. Such design empowers \namemodelnew to detect patterns even if unseen feature nodes are added in the target node. We formally prove this in App.~\ref{app:inductivity}.

\subsubsection{Expressivity}
Expressivity of \gnns is measured by their ability to discriminate non-isomorphic graph structures~\cite{gin,higher_order_wl_gnn} in terms of \textit{$k$-Weisfeiler Leman (WL)} equivalence. As discussed in \S~\ref{sec:specifics}, the Phase-2 of the message-passing layer could adopt any existing \gnn $\Psi$'s message-passing scheme. We show that executing \namemodelnew on the allotropic form $\CG^{alt}$ with $\Psi$ in Phase-2 does not lead to a reduction in expressive power over executing \gnn $\Psi$ on $\CG$. 
\looseness=-1
\begin{thm}[Expressivity of \namemodelnew] \textit{
Let $\Psi_{\CG}\left(v\right): \CV \rightarrow \mathbb{R}^d$ be a trained $L$-layered \gnn on $\CG$.  When an $L$-layered \namemodelnew is trained on $\CG^{alt}$ with $\Psi$ in Phase-2 of message passing to produce $\Psi_{alt}\left(v\right):\CV \rightarrow \mathbb{R}^d$, $\forall \;v_1,v_2 \in \CV$ if $\Psi_{\CG}\left(v_1\right) \neq \Psi_{\CG}\left(v_2\right)$ then representations produced by \namemodelnew are different as well i.e. $\Psi_{\CG}\left(v_1\right) \neq \Psi_{\CG}\left(v_2\right) \rightarrow \Psi_{alt}\left(v_1\right) \neq \Psi_{alt}\left(v_2\right)$ $\forall \; v_1,v_2 \in \CV$.}
\label{thm:expressivity}
\end{thm}
\textbf{Proof:} See App.~\ref{app:expressivity}. $\hfill\square$.
\begin{cor}
\namemodelnew is as expressive as $k$-WL. 
\end{cor}
\textsc{Proof:}  $k$-\gnn~\cite{higher_order_wl_gnn} is as expressive as $k$-WL on static graphs. Therefore, it follows from Thm.~\ref{thm:expressivity}, if $k$-\gnn is used in Phase-2, \namemodelnew is also as expressive as $k$-WL.$\hfill\square$.

Next, we establish that the neural architecture of \namemodelnew is expressive enough to recover the node feature vectors of the original space from the allotropic graph representation.

\begin{thm}
\label{thm:continuousexpressivity}
\textit{ Let $v$ be a node characterized by a $d$-dimensional feature vector $\x_v=[x_1,x_2,\cdots,x_d]$ in the original graph $\CG$. Thus, in the allotropic graph $\CG^{\alt}$, $v$ is connected to $d$ feature nodes with edge weight $x_i$ when connecting to feature node corresponding to dimension $i$. We show that \namemodelnew can recover the original feature vector $\x_v$ from the allotropic graph $\CG^{\alt}$ in Phase 1.}
\end{thm}
\vspace{-0.10in}
\begin{proof} Refer to App.~\ref{app:continuousexpressivity}.
\end{proof}
This result is important since it shows that for Phase 2, \namemodelnew would have the same level of information that its base \gnn would have if operating on the original graph.
\subsubsection{Complexity of \namemodelnew}
In most \gnns such as \graphsage, \gat and \gin, time and space computational complexity for embedding generation of each node is bounded by $O\left(\prod_{\ell=1}^L S_\ell\right)$ \cite{graphsage} where $L$ is no. of layers in \gnn and $S_{\ell}$ is no. of sampled neighbors at each level of the computation graph. In practise, $S_{\ell} \leq K$ is used where $K$ is a small integer. 

In \namemodelnew, this bound increases to $O\left(\prod_{\ell=1}^L \left( S_{\ell}\times{\mid F\mid} + \mid \CV\mid \times \mid F\mid \right)\right)$ due to the $3$-stage message passing layer. In practice, as in \graphsage, one could sample nodes and features to reduce the complexity. We elaborate on these implementation strategies and optimization to exploit sparsity in feature space in App.~\ref{app:complexity}.

\vspace{-0.10in}
\subsection{Continual Learning Framework with \namemodelnew}
\label{sec:continual}
\begin{figure}[t]
\centering
\hspace{-0.2in}
\includegraphics[scale=0.23]{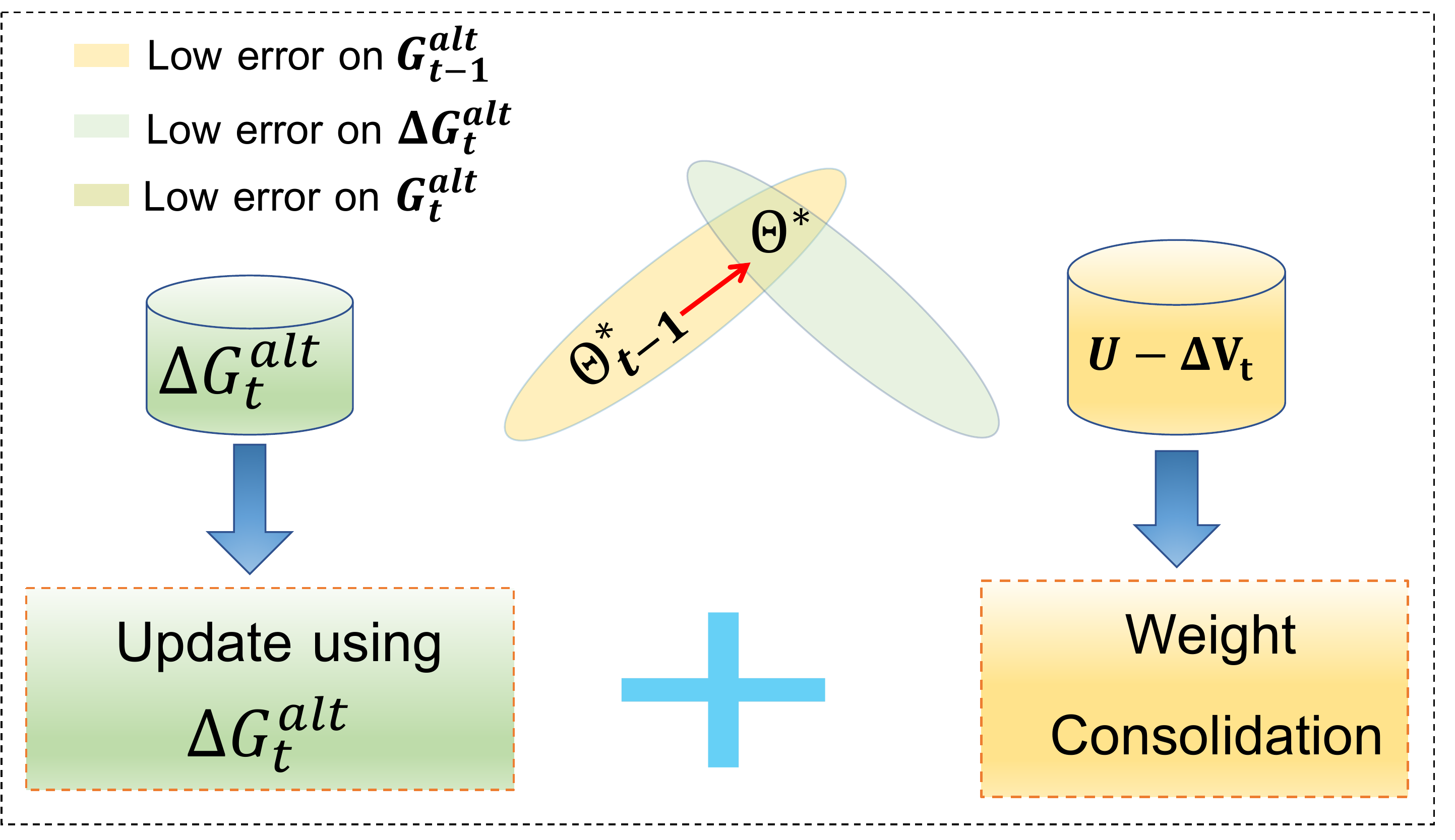}
\caption{Architecture diagram for updating \namemodelnew at time $t$ in a continual fashion. $\Theta_{t-1}^*$ is the optimal parameter space for $\CG^{alt}_{t-1}$. $\Theta^*$ represents the optimal parameter space for $\CG_t^{\alt}$. Since our proposed feature streaming scenario is different from task or class incremental learning, therefore the concept of forgetting in our case is not task-specific but is associated to the unaffected portion of the graph at any given time $t$.}
\label{fig:architecture_cont}
\end{figure}
In this section,  we discuss the adaptation of \namemodelnew to learn on graphs with streaming updates, i.e., $\overrightarrow{\mathcal{\CG}}^{alt}=\left(\CG_1^{alt},\CG_2^{alt}, \ldots \right)$ corresponding to the input streaming graph $\overrightarrow{\mathcal{\CG}}=\left(\CG_1,\CG_2,\ldots\right)$ as described in Def.~\ref{def:streaming_graph}. As the graph is updated, the parameters of \namemodelnew also need to be updated to capture the changes in the graph. 
Our goal is to search for model parameters that fit on the updated portion of the graph while also not forgetting the patterns learned on the unaltered portion. 
Towards that objective, we aim to learn to adjust the magnitude of the parameter updates at time $t$ on certain model weights based on how important they are to the unaffected graph $\CG_t^{alt}-\Delta G_t^{alt}$. We achieve this through \textit{Elastic Weight Consolidation} (EWC)~\cite{continualgnn,kirkpatrick2017overcoming}. Specifically, \namemodelnew penalizes significant changes to parameters that are important for the unaltered graph. Fig.~\ref{fig:architecture_cont} illustrates  \textit{Elastic Weight Consolidation} for our streaming feature scenario. The specifics of this component are discussed in App.\ref{app:continual}.

\section{Experiments}
\vspace{-0.05in}
In this section, we examine the effectiveness of  \namemodelnew wrt. \textbf{(1)} Robustness to different missing feature rates \textbf{(2)} Adaptation to different  \gnn architectures \textbf{(3)} Ablation study, and \textbf{(4)} Performance on continual learning. Details of the experimental setup in terms of hardware and software framework, train-test splits, default parameter value, etc., are listed in App.~\ref{app:environment}. Our codebase  is available at \url{https://github.com/data-iitd/Grafenne}.

\subsection{Datasets}
\vspace{-0.05in}
We evaluate \namemodelnew on the real-world graphs listed in Table~\ref{tab:dataset_details}. Among these, Actor is a heterophilic graph, whereas the rest are homophilic.
Further details on the semantics of the datasets are provided in App.~\ref{app:datasets}.  
\begin{table}[b]
\caption{Dataset statistics}
\vspace{-0.10in}
\centering
\scriptsize
\begin{tabular}{p{2.8cm}cccc}
\toprule
\textbf{Dataset} & \textbf{\# Nodes} & \textbf{\# Edges} & \textbf{\# Features} & \textbf{\#Labels} \\
\midrule
Cora~\cite{cora} & $2708$ & $10556$ & $1433$ & $7$ \\ 
CiteSeer~\cite{sslge} & $3327$ & $9104$ & $3703$ & $6$ \\ 
Physics~\cite{shchur2018pitfalls} & $34493$ & $495924$ & $8415$ & $5$ \\
Actor~\cite{geomgcn} & $7600$ & $33544$ & $931$ & $5$ \\ 
\bottomrule
\end{tabular}
\label{tab:dataset_details}
\end{table}

\subsection{Baselines}
To deal with missing features, we consider  five different feature imputation strategies namely: \textbf{(1)} \gcnmf~\cite{gcnmf}, \textbf{(2)} \pagnn~\cite{pagnn}, \textbf{(3)} \fp~\cite{fp}, \textbf{(4)} Marking missing features with a special label, \textbf{(5)} Imputing based on the mean of the neighborhood values. Among the above strategies, \gcnmf and \pagnn propose their own \gnns. In contrast, the other three algorithms are all pre-processing methods and therefore can be integrated with any \gnn of choice. As the base \gnn architecture, we consider \graphsage~\cite{graphsage}, \gin~\cite{gin}, \gat~\cite{gat}. \gcnmf, \pagnn, and \fp are all transductive, while the other two are inductive. 
{We also compare with ~\fate~\cite{wu2021towards}, which is an algorithm for feature adaptation. We show that feature adaptation methods, when adapted for graphs, are not adequate. A detailed differentiation in methodology and empirical comparison is provided in App.~\ref{app:comp_FATE}.}

For the pre-processing based methods, we use the notations ``\fp+$\langle\gnn\rangle$'' and ``\nm+$\langle\gnn\rangle$'' for \fp and neighborhood mean respectively. If we only use the name of the \gnn, then it indicates imputation with a special label for missing value. 
\looseness=-1

\subsection{Tasks}
\label{sec:tasks}
\namemodelnew is generic enough to accommodate any of the standard predictive tasks on graphs. We choose two of the most popular tasks of \textit{node classification} and \textit{link prediction} to benchmark \namemodelnew and the baselines. As per standard practice~\cite{graphsage}, for node classification, we quantify performance in terms of accuracy, i.e., the percentage of correct predictions, and for link prediction, we use area under the receiver operating curve (AUCROC).

\subsection{Empirical Evaluation}
\label{exp:emipirical_eval}
First, we evaluate on static graphs with heterogeneous features sets. Next, we evaluate performance on streaming graphs. Finally, we perform ablation studies. Since all of the pre-processing features require a base \gnn, we primarily use \graphsage as the \gnn of choice. To ensure a fair comparison, we also use \graphsage as the message passing scheme in Phase-2 of \namemodelnew. Nonetheless, for the sake of completeness, we also present results when \graphsage is replaced with \gat and \gin. To measure the impact of missing features, we take datasets with complete features, and randomly delete $p$ portion (ratio) of the features per node. $p$ is varied across various values. This strategy is consistent with evaluation methodology of our baselines \pagnn,\gcnmf, and \fp.
\begin{table}[t]
\caption{Accuracy of \namemodelnew and baselines on node classification at various missing rates $p$. Std. dev. values $<$ $0.01$ are approximated to $0$.}
\centering
\resizebox{\columnwidth}{!}{%
\begin{tabular}{llccc}

\toprule
\textbf{Dataset} & \textbf{Method} & \boldmath{$p=0$} &\boldmath{$p=0.5$} &  \boldmath{$p=0.99$}\\ 
\midrule
\multirow{8}{*}{Cora} & \graphsage &$83.80 \pm 0.48$ & $83.06 \pm 0.62$  &  $72.58 \pm 0.71$ \\ 
 & \gcnmf & $80.07 \pm 0.0$ &$67.52 \pm 0.0$   & $33.02 \pm 0.0$ \\
 & \pagnn & $82.47 \pm 0.0$ &$84.68 \pm 0.0$  & $67.89 \pm 0.0$ \\
 &\nm+ \graphsage & - &$83.46 \pm 0.36$  & $78.48 \pm 0.54$ \\
 &\fp+ \graphsage & - &$83.72 \pm 0.53$  & $81.25 \pm 0.44$ \\
 \cmidrule{2-5}
 &\namemodelnew & \boldmath{$87.6 \pm 0.73$} &$84.35 \pm 0.27$  & $78.85 \pm 0.29$ \\ 

 &\nm+ \namemodelnew & - & $85.05 \pm 0.36$  & $78.78 \pm 0.62$ \\
 & \fp+ \namemodelnew & - &\boldmath{$85.46 \pm 0.21$} &  \boldmath{$82.91 \pm 0.92$} \\
\cmidrule{1-5}

\multirow{8}{*}{CiteSeer} & \graphsage &$73.43 \pm 0.97$ & $70.85 \pm 0.35$   & $57.14 \pm 0.96$ \\ 
 & \gcnmf & $71.47 \pm 0.0$ &$60.36 \pm 0.0$   & $23.12 \pm 0.0$ \\
 & \pagnn & $73.57 \pm 0.0$ &$72.82 \pm 0.0$   & $58.70 \pm 0.0$ \\
 &\nm+ \graphsage & - &$70.96 \pm 0.45$  & $61.80 \pm 0.46$ \\
 &\fp+ \graphsage & - &$71.02 \pm 0.65$  & \boldmath{$65.25 \pm 1.08$} \\
 \cmidrule{2-5}
 &\namemodelnew & \boldmath{$73.90 \pm 0.84$} &$72.91 \pm 0.95$   & $64.08 \pm 0.79$ \\ 

 &\nm+ \namemodelnew & - & $72.88 \pm 0.55$ &  $63.03 \pm 0.93$ \\
 & \fp+ \namemodelnew & - &\boldmath{$74.20 \pm 0.40$}  & {$64.64 \pm 0.6$} \\
\cmidrule{1-5}

\multirow{8}{*}{Actor} & \graphsage &$32.90 \pm 0.79$ & $30.61 \pm 0.86$  & $22.90 \pm 0.50$ \\ 
 & \gcnmf & $24.53 \pm 0.0$ &$23.75 \pm 0.0$   & $21.57 \pm 0.0$ \\
 & \pagnn & $23.81 \pm 0.0$ &$23.81 \pm 0.0$   & \boldmath{$25.39 \pm 0.0$} \\
 &\nm+ \graphsage & - &$29.27 \pm 0.69$  & $21.73 \pm 0.07$ \\
 &\fp+ \graphsage & - &$29.15 \pm 0.79$  & $23.89 \pm 1.03$ \\
 \cmidrule{2-5}
 &\namemodelnew & \boldmath{$38.90 \pm 0.84$} &\boldmath{$35.02 \pm 0.21$}   & $23.97 \pm 0.58$ \\ 

 &\nm+ \namemodelnew & - & $32.03 \pm 0.70$  & $24.01 \pm 0.80$ \\
 & \fp+ \namemodelnew & - &$32.76 \pm 1.06$  & $24.02 \pm 0.40$ \\



\bottomrule

\end{tabular}%
}
\label{tab:res:Gsage}
\end{table}
\begin{table}[bh!]
\caption{Accuracy of \namemodelnew(\gat) and \namemodelnew(\gin) with benchmark \gnns, \gat and \gin. }
\centering
\resizebox{\columnwidth}{!}{%
\begin{tabular}{llccc}

\toprule
\textbf{Dataset} & \textbf{Method} & \boldmath{$p=0$} & \boldmath{$p=0.5$}  & \boldmath{$p=0.99$}\\ 
\midrule
\multirow{4}{*}{Cora} & \gat & \boldmath{$86.10 \pm0.7$} & $82.50 \pm0.96$  & $73.72 \pm0.57$\\
 & \namemodelnew(\gat) & $85.97 \pm0.53$ & \boldmath{$85.16 \pm0.63$}  & \boldmath{$79.74 \pm0.51$}\\
\cmidrule{2-5}

 & \gin & $85.09 \pm0.92$ & $82.91 \pm0.89$ &  $73.28 \pm0.35$\\
 & \namemodelnew(\gin) & \boldmath{$85.94 \pm0.39$} & \boldmath{$84.25 \pm0.64$} & \boldmath{$82.36 \pm1.17$}\\
\cmidrule{1-5}

\multirow{4}{*}{CiteSeer} & \gat & $71.59 \pm0.85$ & $69.15 \pm0.91$  & $59.21 \pm0.73$\\
 & \namemodelnew(\gat) & \boldmath{$73.21 \pm0.33$} & \boldmath{$72.64 \pm0.76$}  & \boldmath{$64.29 \pm0.61$}\\
\cmidrule{2-5}

& \gin & $72.16 \pm0.58$ & $69.84 \pm1.10$  & $60.15 \pm1.31$\\
 & \namemodelnew(\gin) & \boldmath{$73.45 \pm1.04$} & \boldmath{$72.58 \pm0.59$}  & \boldmath{$64.32 \pm1.15$}\\
\cmidrule{1-5}

\multirow{4}{*}{Actor} & \gat & $26.68 \pm0.92$ & $25.92 \pm0.49$ & \boldmath{$24.69 \pm1.55$}\\
 & \namemodelnew(\gat) & \boldmath{$33.86 \pm0.88$} & \boldmath{$32.07 \pm0.67$}  & $24.23 \pm0.4$\\
\cmidrule{2-5}

& \gin & $26.93 \pm0.76$ & $26.48 \pm1.36$  & $22.98 \pm0.56$\\
 & \namemodelnew(\gin) & \boldmath{$29.13 \pm0.91$} & \boldmath{$29.15 \pm1.36$}  & \boldmath{$24.03 \pm0.5$}\\



\bottomrule

\end{tabular}%
}
\label{tab:res:GATGIN} 
\end{table}
\vspace{-0.10in}
\subsubsection{Static graphs}
Table~\ref{tab:res:Gsage} show the results on node classification at multiple feature missing rates. A similar table for link prediction is provided in Table~\ref{tab:res:link_pred_baselines}. We observe that \namemodelnew outperforms baseline methods on a diverse range of missing rates. Especially on higher missing rates, we observe a performance gap of more than $10\%$ between \namemodelnew and the best baseline method. Further, on dataset Actor, we observe that \namemodelnew obtains significantly better accuracy gain of over $5\%$ on all missing rates. Interestingly, we also observe that the performance of \fp and mean-neighborhood (\nm) methods can be significantly improved when complemented with \namemodelnew's message passing framework, i.e., \nm+\namemodelnew and \fp+\namemodelnew. We note that \fp is a transductive feature imputation method that requires re-training in cases of unseen nodes or new features, making \namemodelnew an attractive alternative in streaming graphs. These results are a direct consequence of the nature of propagation introduced by our proposed method. We also evaluate \namemodelnew on extremely high missing rates such as $p=0.99999$ on large-scale dataset physics in table \ref{tab:large_scale_missing} where see \namemodelnew performs exceptionally well indicating its application in settings where the nominal amount of data is shared by very few users. 

In Tables~\ref{tab:res:GATGIN} and \ref{tab:res:link_pred} (in appendix), we investigate the impact of the \gnn used in Phase-2 of \namemodelnew's message passing scheme. Towards that, \graphsage is replaced with \gat and \gin. We observe that regardless of the \gnn, when empowered within the \namemodelnew framework, an improvement is observed. Interestingly, even when almost all features are available ($p=0$), for the majority of the cases, \namemodelnew outperforms solely using a \gnn on the original graph.

\begin{table}[h]
\caption{AUCROC of \namemodelnew and baselines on link prediction task . Note that we have not included \gcnmf and \pagnn as their code adapted for the link prediction task is not generalizing on test data }
\centering
\resizebox{\columnwidth}{!}{%
\begin{tabular}{llccc}

\toprule
\textbf{Dataset} & \textbf{Method} & \boldmath{$p=0$} & \boldmath{$p=0.5$} & \boldmath{$p=0.99$}\\ 
\midrule

\multirow{6}{*}{Cora} & \graphsage & {$0.86 \pm0.002$} & $0.84 \pm0.002$ &  $0.7523 \pm0.05$\\
 & \nm+~\graphsage & - & $0.8687 \pm0.0008$   & $0.8297 \pm0.0014$\\
 & \fp+~\graphsage & - & $0.8773 \pm0.0015$   & $0.9137 \pm0.0008$\\
 \cmidrule{2-5}
& \namemodelnew & \boldmath{$0.8780 \pm0.004$} & {$0.8501 \pm0.005$}   & {$0.8015 \pm0.006$}\\
& \nm+~\namemodelnew & - & $0.9009 \pm0.0026$   & $0.8632 \pm0.0007$\\
& \fp+~\namemodelnew & - & \boldmath{$0.9344 \pm0.0020$}   & \boldmath{$0.9263 \pm0.0012$}\\


\cmidrule{1-5}

\multirow{6}{*}{CiteSeer} & \graphsage & {$0.8251 \pm0.005$} & $0.7617 \pm0.001$ &  $0.7223 \pm0.002$\\
 & \nm+~\graphsage & - & $0.8213 \pm0.0018$   & $0.8001 \pm0.0019$\\
 & \fp+~\graphsage & - & $0.8506 \pm0.0014$   & $0.8875 \pm0.0011$\\
 \cmidrule{2-5}
& \namemodelnew & \boldmath{$0.8681 \pm0.0010$} & {$0.8047 \pm0.010$}   & {$0.7378\pm0.008$} \\
& \nm+~\namemodelnew & - & $0.8944 \pm0.0067$   & $0.8462 \pm0.0024$\\
& \fp+~\namemodelnew & - & \boldmath{$0.9348 \pm0.0024$}   & \boldmath{$0.9012 \pm0.0030$}\\

\cmidrule{1-5}

\multirow{6}{*}{Actor} & \graphsage & {$0.6569 \pm0.0013$} & $0.7029 \pm0.0026$ &  $0.6969 \pm0.003$\\
 & \nm+~\graphsage & - & $0.7473 \pm0.0004$   & $0.6885 \pm0.0001$\\
 & \fp+~\graphsage & - & $0.7552 \pm0.0003$   & $0.7721 \pm0.0017$\\
 \cmidrule{2-5}
& \namemodelnew & \boldmath{$0.7047 \pm0.0050$} & {$0.7021 \pm0.005$}   & {$0.7029\pm0.007$}\\
& \nm+~\namemodelnew & - & $0.7562 \pm0.0032$   & $0.7169 \pm0.0012$\\
& \fp+~\namemodelnew & - & \boldmath{$0.7801 \pm0.0026$}   & \boldmath{$0.7869 \pm0.0013$}\\

\bottomrule
\end{tabular}%
}
\label{tab:res:link_pred_baselines} 
\end{table}

\begin{table*}[t!]
\caption{Node classification performance comparison at extreme missing rates $p$ on large-scale Physics dataset. We report the mean classification accuracy (\%) along with the standard deviation on five runs. Std. dev. values $<$ $0.01$ are approximated to $0$. If a baseline produces an error during execution, we denote it as $*$.}
\vspace{-0.1in}
\centering
\resizebox{0.9\textwidth}{!}{%
\begin{tabular}{lcccccc}

\toprule
 \textbf{Method} & \boldmath{$p=0$} & \boldmath{$p=0.5$} & \boldmath{$p=0.99$} & \boldmath{$p=0.999$} & \boldmath{$p=0.9999$} & \boldmath{$p=0.99999$}\\ 
\midrule
 \graphsage & $96.91 \pm0.05$ & $96.29 \pm0.17$ & $92.92 \pm0.11$ & $84.11 \pm0.08$ & $61.05 \pm0.27$ & $52.11 \pm0.04$\\
  \gcnmf & $92.52 \pm0.0$ & {$81.12 \pm0.0$} & $50.4856 \pm0.0$ & $51.45 \pm 0.0$ & $*$ & $*$\\
 \pagnn & $94.28 \pm0.0$ & {$93.88 \pm0.0$} & $88.09 \pm0.0$ & $75.44 \pm 0.0$ & $60.71 \pm 0.0$ & $51.35 \pm 0.0$\\
 
 \nm+~\graphsage & $-$ & {$96.08 \pm0.11$} & $94.47 \pm0.14$ & $92.35 \pm 0.06$ & $82.96 \pm 0.07$ & $57.48 \pm 0.01$\\
     \fp+~\graphsage & $-$ & {$96.41 \pm0.09$} & $95.04 \pm{0.0}$ & $94.34 \pm 0.15$ & $93.07 \pm 0.22$ & $78.44 \pm 0.71$\\

\cmidrule{1-7}
  \namemodelnew & \boldmath{$97.02 \pm0.05$} & {$96.23 \pm0.23$} & $94.49 \pm0.18$ & $94.31 \pm0.19$& $94.11 \pm0.21$& $89.49 \pm0.15$\\
\nm +~\namemodelnew & $-$ & {$95.82 \pm0.12$} & $94.70 \pm0.18$ & ${\mathbf{94.61} \pm\mathbf{0.0}}$& $94.57 \pm0.19$& $89.56 \pm0.06$\\
\fp +~\namemodelnew & $-$ & {${\mathbf{96.36} \pm\mathbf{0.08}}$} & ${\mathbf{95.02} \pm\mathbf{0.05}}$ & $94.57 \pm0.15$& \boldmath{$95.33 \pm0.21$} & \boldmath{$93.17 \pm0.16$}\\




\bottomrule

\end{tabular}%
}
\label{tab:large_scale_missing} 
\end{table*}

\begin{figure*}[h!]
 \vspace{-0.2in}
\centering
\subfloat[Physics]{
\includegraphics[width=0.405\textwidth]{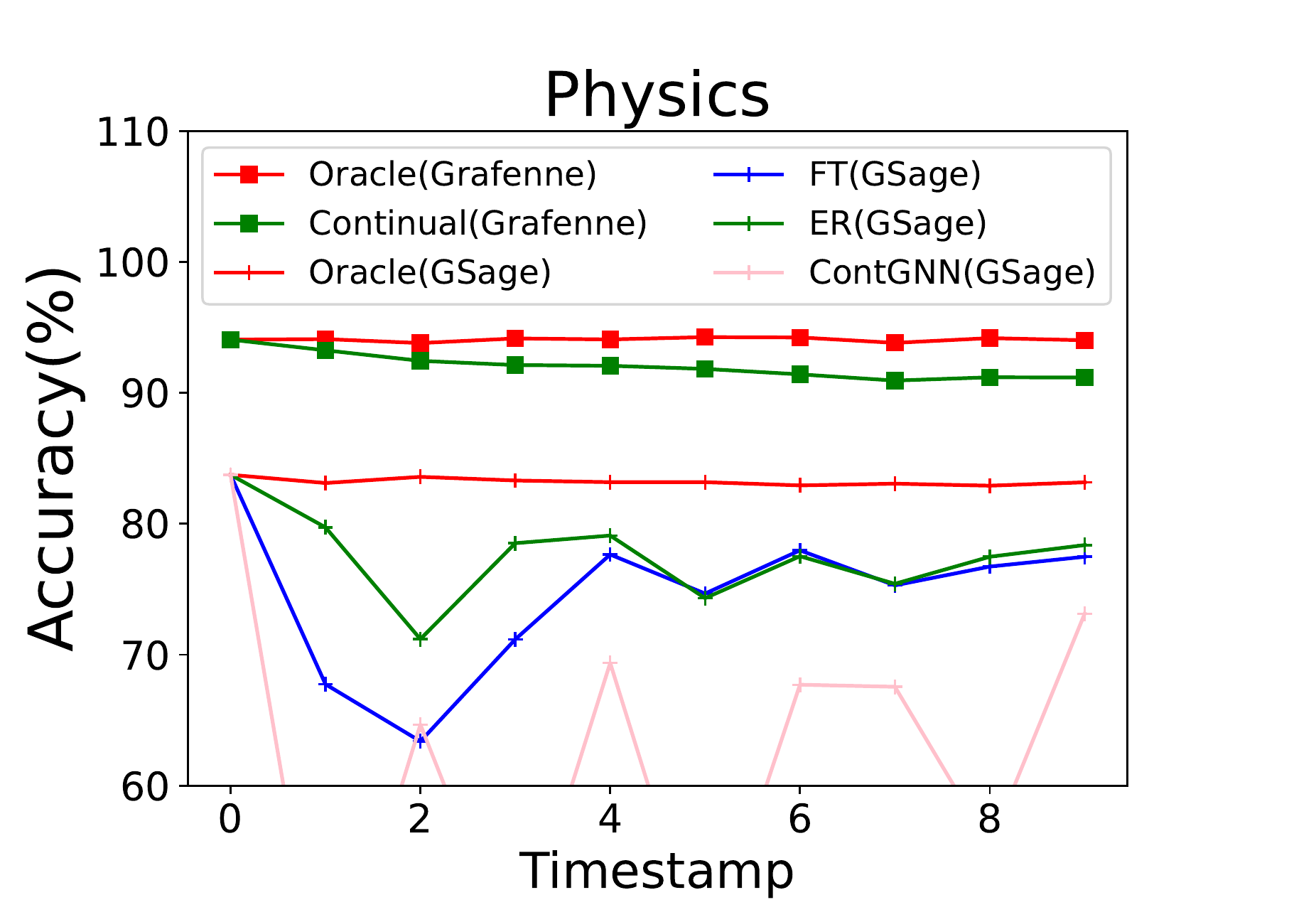 }
\label{fig:phycontinual}
  }
\hspace{-0.3in} 
\subfloat[Cora]{
\includegraphics[width=0.35\textwidth]{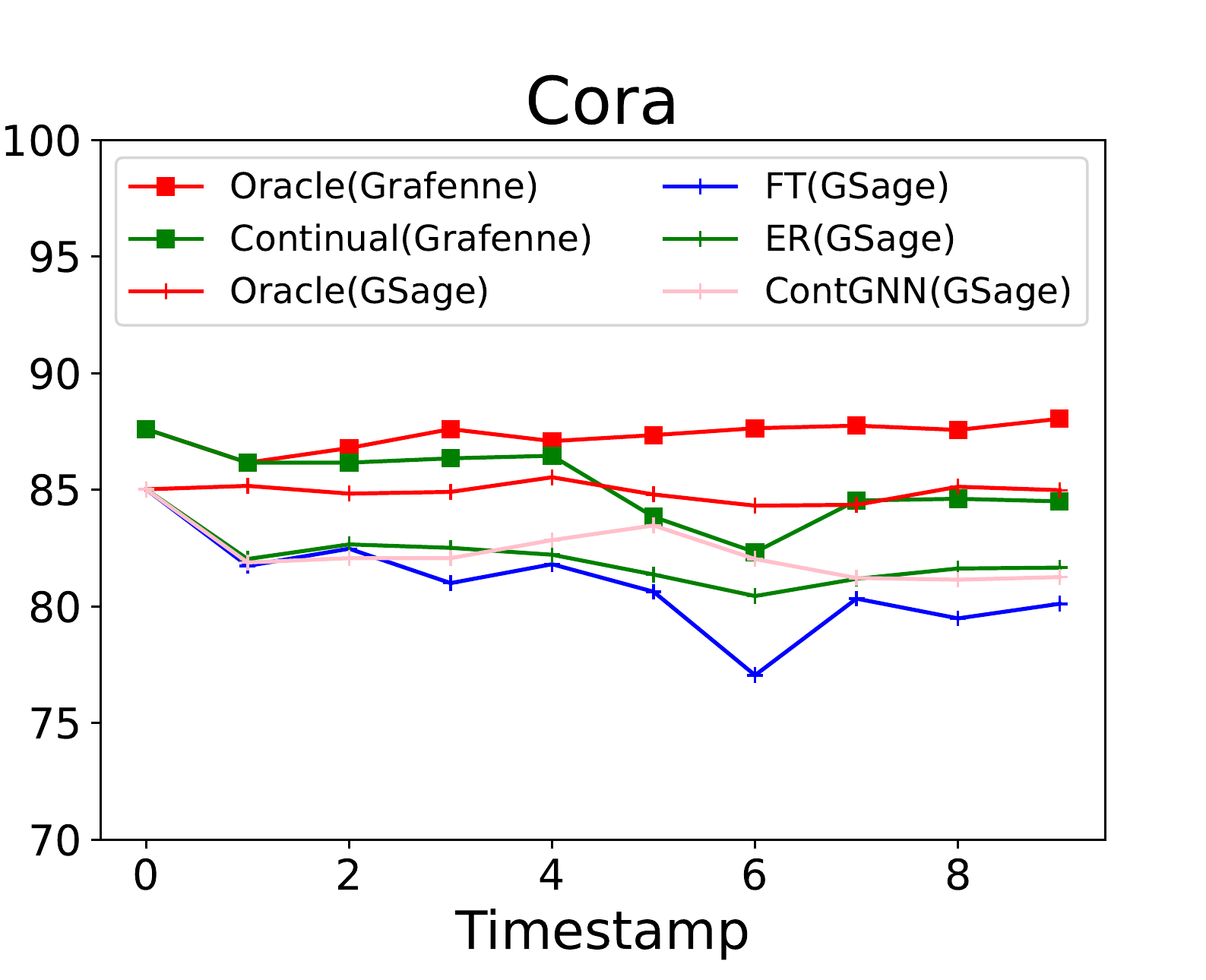 }
\label{fig:citeseercontinual}
  }
\vspace{-0.15in} 
\hspace{-0.35in} \\
\subfloat[CiteSeer]{   \hspace{-0.12in}
  \includegraphics[width=0.405\textwidth]{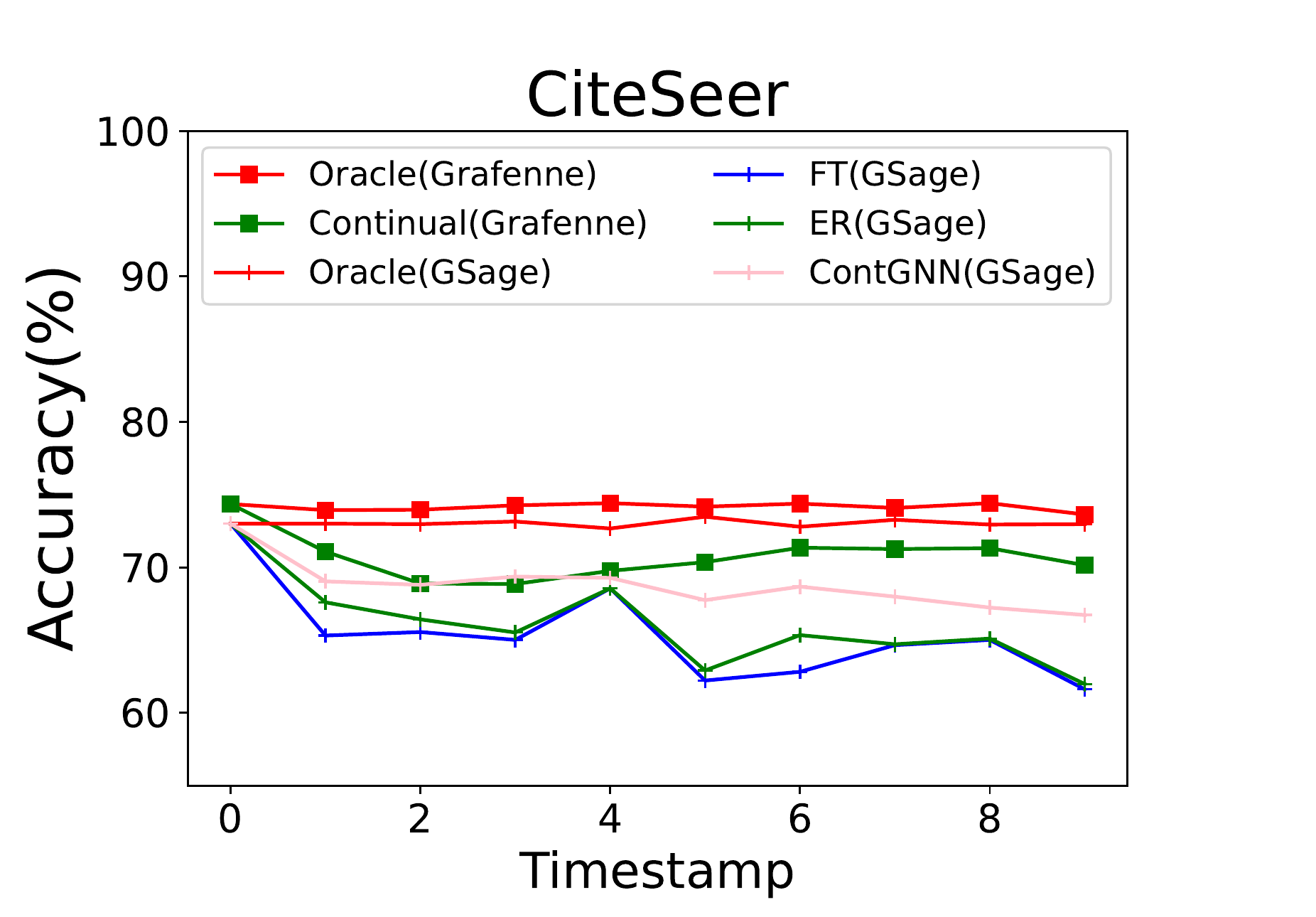} 
  }\hspace{-0.12in} 
  \subfloat[DBLP]{ \hspace{-0.15in} 
  \includegraphics[width=0.35\textwidth]{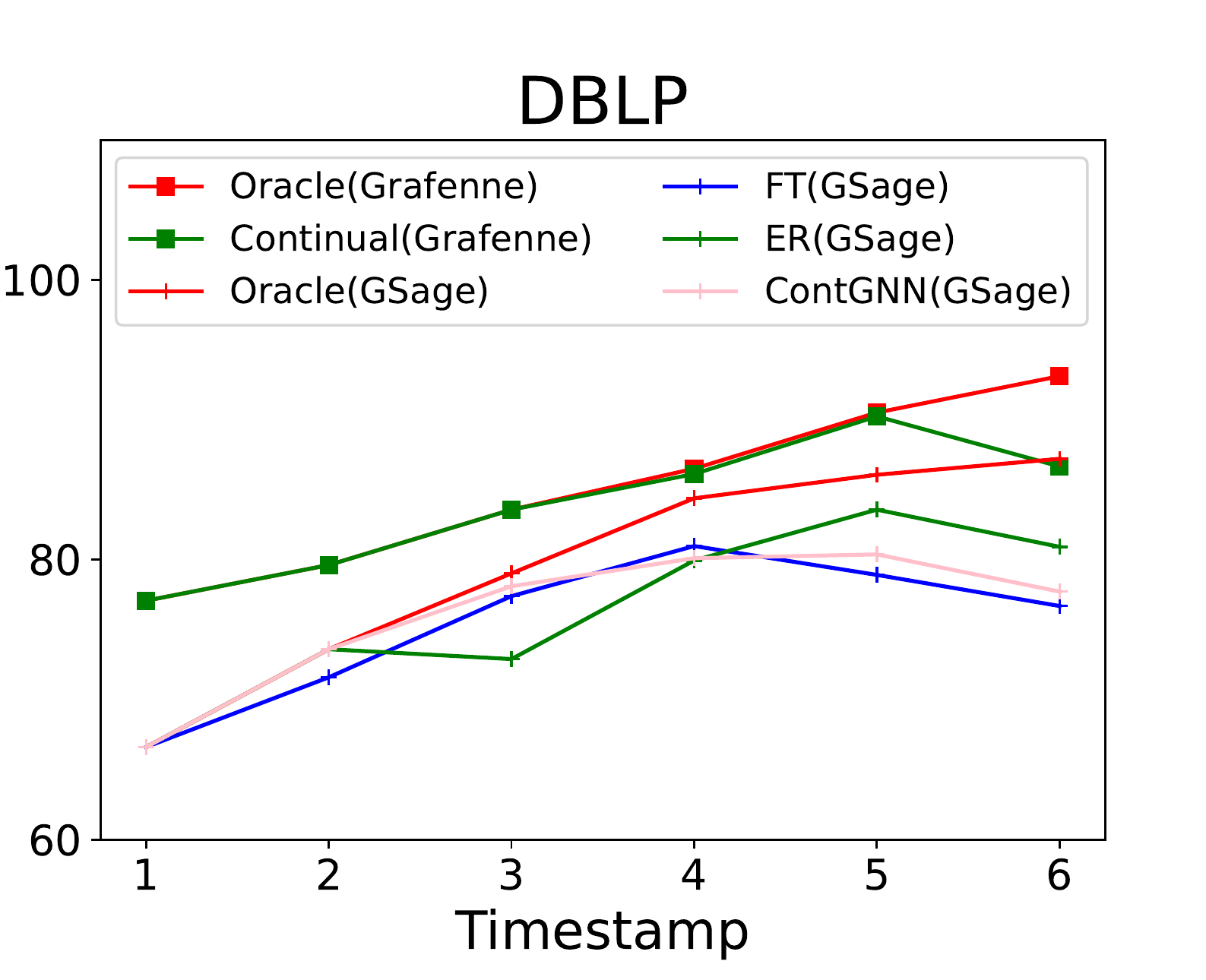} 
  }
\vspace{-0.1in}
 \caption{ Continual learning performance on Physics, Cora,  CiteSeer and DBLP. The $x$-axis represents the  timestamps of graph updates and the $y$-axis represents the test accuracy(\%) corresponding to each timestamp.}
\label{fig:continualplots}
\end{figure*}

\subsubsection{Continual Learning with \namemodelnew}

\noindent
\textbf{Setup:}
To evaluate \namemodelnew for continual learning, we evaluate on dynamic graphs, where features get added/deleted over time and graph structure also changes over time.
\vspace{-0.10in}
\begin{itemize}

\item \textbf{Feature Addition and Deletion:} A subset of features for a subset of nodes is added or deleted at each timestamp. We first sample nodes with probability $p_n$. For each selected node, we randomly select features for addition/deletion. The probability of a feature getting added or deleted to a node is $p_{f}^{add}$ and $p_{f}^{del}$ respectively. 

\item \textbf{Edges:} A subset of edges in the graph are added/deleted over time. The probability of an edge getting selected for deletion is $p_{e}^{del}$ and the probability of an edge getting added to the graph is  $p_{e}^{add}$.
\end{itemize}
In our experiments, for Cora and CiteSeer we set $p_n=0.03$,  $p_{f}^{add}=0.05, p_f^{del}=0.4, p_e^{del} =p_e^{add} =0.0005$ and $T=9$. For Physics dataset we set   $p_n=0.003, p_f^{del}=0.8, p_f^{add}=0.0001, p_e^{del} =p_e^{add} =0.00005$ . 

\textit{Real-world dynamic dataset:} Additionally, we extract streaming DBLP dataset~\cite{tang2008arnetminer}
 between 1992 to 1997 where nodes and edges get added over time. This dataset has  $3075$ nodes, $6368$ edges and  $5$ classes. We set  $p_n=0.05, p_f^{del}=0.1, p_f^{add}=0.05$. 
 
We set $\lambda$ defined in Eq.~\ref{eq:continual_update} to $100000$. We set $|U|=300$ for DBLP and $|U|=25$ for other datasets(\S~\ref{sec:continual}). 

\noindent\textbf{Baselines:} Given a graph stream $\overrightarrow{\CG}=(\CG_1,\CG_2 \ldots, \CG_T)$, we compare \namemodelnew for continual learning with: \\
    \noindent \textbf{(1) \textsc{Oracle}: Retraining from scratch:} 
    \textsc{Oracle} discards existing parameters  and retrains \namemodelnew from scratch on $\CG_t$ resulting in optimal parameters $\Theta_t$ for $\CG_t$. This method provides the upper bound of achievable performance.\\
    \noindent \textbf{(2) FT: Fine-tuning:} We update the parameters using \textit{only} the affected nodes in $\CG_t$, i.e., $\Delta \CV_t$. \\
    \textbf{(3) ER: Experience Replay}: Here, in addition to FT, we preserve a small sample of past nodes in memory and replay them when training on $\CG_{t}$ to avoid forgetting of past patterns. \\
    \textbf{(4)} \textbf{ContGNN}\cite{wang2020streaming}: Preserves past knowledge through a combination of experience replay and weight regularization.



\noindent
\textbf{Results on Continual Learning Scenario:}
In Fig.~\ref{fig:continualplots}  we compare the performance of proposed continual method for training \namemodelnew along with Oracle,  FT, ER and ContGNN\cite{wang2020streaming} methods. We report test performance on the entire graph at each timestamp for every method. In Fig. \ref{fig:continualplots}, we observe that the continual method on \namemodelnew can maintain significantly higher accuracy on the entire test-set compared to other methods.  The fine-tuning method only updates the affected portion of the graph, hence it suffers from catastrophic forgetting on the unaffected nodes, which is evident from the accuracy metric on the test nodes from the whole graph. Methods such  as ER and ContGNN preserve past knowledge by employing a small set of memory for replay. However, they do not cater to the unseen feature scenario, hence overall performance on these methods deteriorates over time. On the other hand,   \namemodelnew  coupled with elastic weight consolidation on an  unaffected portion of graph reduces the extent of catastrophic forgetting, hence achieves superior performance. Additionally, in the case of large-size datasets i.e Physics, we observe that \namemodelnew significantly outperforms existing methods on all timestamps showing better scalability.

\section{Conclusion}
Graph Neural Networks have shown significant performance gains on graph-structured data. However existing works mostly focused on graphs with an identical set of available node features. Moreover, existing state-of-the-art imputation techniques to tackle scenario of  dissimilar node features are transductive in nature. In this work, we proposed a novel inductive method \namemodelnew that can learn on graphs having nodes with heterogeneous features. In addition to this, we also formulated a novel problem of lifelong learning on graphs with streaming features. Further, to solve this problem, we proposed \textit{elastic weight consolidation} based continual learning method for training \namemodelnew on dynamic graphs.  Through extensive evaluation on 4 real-world datasets, we established that \namemodelnew achieves superior performance against baseline approaches at various feature missing rates $p$ and is  also robust  at extremely high missing feature rates e.g. $p=0.99999$. Furthermore, we highlight the capability of \namemodelnew to integrate with different existing inductive \gnn architectures and show significant performance gains. Additionally, \namemodelnew achieves high-quality results in the streaming scenario and hence shows its ability to learn effectively in the lifelong learning setup. In terms of future work, it will be interesting to explore \namemodelnew on graphs having inter-feature relations allowing the creation of feature-feature edges in $\CG^{\alt}$. 
\looseness=-1


\section{Acknowledgement}
Shubham Gupta acknowledges Info Edge (India) Limited for supporting his Ph.D. Sahil Manchanda acknowledges Qualcomm for supporting him through Qualcomm Innovation Fellowship and he also acknowledges GP Goyal Alumni Grant of IIT Delhi for supporting this travel. Sayan Ranu acknowledges the Nick McKeown chair position endowment. Srikanta Bedathur was partially supported by DS Chair Professor of AI grant and an IBM AI Horizons Network (AIHN) grant. 


\bibliography{references}
\bibliographystyle{icml2023}

\newpage
\appendix
\onecolumn

\section{Appendix}
\label{sec:appendix}

\appendix
\renewcommand{\thesubsection}{\Alph{subsection}}
\renewcommand{\thefigure}{\Alph{figure}}
\renewcommand{\thetable}{\Alph{table}}

\subsection{Phase 2 Specifics of Message Passing}
\label{app:phase2}
Phase 2 is a message-passing layer among graph nodes $\CV$. \namemodelnew adopts the message passing implementation of \graphsage, which is defined as follows.
\noindent

\textbf{\namemodelnew:}\vspace{-0.04in}
\begin{equation}
    \h_v^{\ell} = \sigma \left({\W^{\ell}}_{13}^T\left(\h_v^{l} \bigparallel \frac{1}{\mid \mathcal{N}_v^{feat}\mid} \sum_{u \in \mathcal{N}_v^{feat}} \h_u^{l} \right) \right) \quad  \forall v \in \CV
\label{eq:fognn_gsage}
\end{equation}

The message passing layer of \namemodelnew is flexible enough to accommodate other architectures as well such as \textsc{Gat}~\cite{gat} and \textsc{Gin}~\cite{gin}.

\noindent\textbf{\namemodelnew(\gat):} $\forall v \in \CV$
\begin{equation}
\begin{gathered}
    \m_v^{\ell}\left(u\right){=}\text{\textsc{LeakyReLU}}\left({\W_{13}^{\ell}}\h_v^{l}{ \bigparallel }{\W_{14}^{\ell}}\h_u^{l}\right)\;\; \forall  u {\in }\mathcal{N}_v^{\CG} {\cup }\;v,\\
    \alpha_{vu}{ =} \frac{\exp\left({\w_{15}^{\ell}}^T\m^{\ell}_v\left(u\right)\right)}{\sum\limits_{u' \in \mathcal{N}_{v}^{\CG}{ \cup}\; v }\exp\left({\w^{\ell}_{15}}^T\m^{\ell}_v\left(u'\right)\right)},\;\\ \h_v^{\ell}\;\;{ = }{\sum_{u \in \mathcal{N}_{v}^{\CG}{ \cup}v}}{ \alpha_{vu}\W^{\ell}_{16}\h_u^{\ell}}
\end{gathered}
\label{eq:fognn_gat}
\end{equation}
\noindent\textbf{\namemodelnew(\gin):}
\begin{equation}
    \h_v^{\ell} = \text{\textsc{MLP}}\left(\left(1+\epsilon\right)\h_v^{\ell} + \sum_{u \in \mathcal{N}^G\left(v\right)}\h_u^{\ell}\right) \quad  \forall v \in \CV
\label{eq:fognn_gin}
\end{equation}
All weights matrices and vectors   of the form $\W^{\ell}_i$ and $\w^{\ell}_i$ are trainable parameters; $\epsilon$ is a hyper-parameter.

\subsection{Inductive Analysis of \namemodelnew}
\label{app:inductivity}
We first define an input test graph $\CG_{test}=(\CV_{test},{\CE}_{test},\X_{test})$, which is an updated graph of input training graph $\CG$ i.e. $\CG_{test}=\CG+\nabla \CG$ where $\nabla \CG= (\nabla {\CV},\nabla {\CE},\nabla \X)$ . Similar to def. \ref{def:graph}, where we define $F$ as set of available features in graph $\CG$, we also define feature set $F_{test}=\bigcup _{v \in \CV_{test}}F_v $ on $G_{test}$ where $F_v$ is set of features available at node $v$. We also assume an unknown feature super-set $\mathcal{F}$ where $F, F_{test}\subseteq \mathcal{F}$. Now, in lieu of proposition \ref{props:inductivity}, we describe and prove the following theorem \ref{thm:inductivity}. 
\looseness=-1
\begin{thm}[Inductivity of \namemodelnew] \textit{
Let $\Psi_{\CG}\left(v\right): \CV \rightarrow \mathbb{R}^d$ be a trained $L$-layered \gnn on a graph $\CG$ and $\Psi_{alt}\left(v\right):\CV \rightarrow \mathbb{R}^d$ be a $L$-layered \namemodelnew trained on $\CG^{alt}$ with $\Psi$ in Phase-2 of message passing. Given a test graph $\CG_{test}$, \namemodelnew $\Psi_{alt}$ can generalize to unseen nodes with unseen features in $\CG_{test}^{alt}$ i.e. $\Psi_{alt}(v) \approx Y(v) $ even $if\;\exists f \in F_v \wedge f \notin F,\;\;\forall v \in {\CV}_{test}-{\CV}$ . This holds true given that the following conditions hold.
\begin{enumerate}
    \item $\Psi$ is a node-inductive \gnn i.e. it is able to generalize to unseen nodes albeit with seen features.
    \item A feature embedding space $\mathcal{Z} \subseteq \mathbb{R}^d$ which reflects the semantic/statistical relationships among vectors and an embedding function over categorical variables $\Phi(f):\mathcal{F}\rightarrow \mathcal{Z}$ which can map any seen or unseen feature to this embedding space $\mathcal{Z}$ i.e., $\Phi(f) \in \mathcal{Z},\;\forall f \in F_{test} - F$.
\end{enumerate}
}
\label{thm:inductivity}
\end{thm}
\textbf{Proof:} Condition-1 always holds, as \namemodelnew adopts the node inductive \gnn $\Psi$ in Step-2 as seen in eq. \ref{eq:fognn_gnode_gnode_msg}, \ref{eq:fognn_gnode_msg_agg} and \ref{eq:fognn_gnode_combine} which are independent of no. of graph nodes $|\CV|$ in $G^{alt}$. All graph nodes $\CV$ are initialized with $\boldsymbol{0}$ and assuming that condition-2 holds true, all feature nodes $\CV^{feat}$ are initialized with vectors in embedding space $\mathcal{Z}$. Combining this and the fact that eq. \ref{eq:fognn_feat_msg}, \ref{eq:fognn_feat_msg_agg}, \ref{eq:fognn_feat_combine}, \ref{eq:fognn_gnode_feat_msg}, \ref{eq:fognn_gnode_feat_msg_agg} and \ref{eq:fognn_gnode_feat_combine} are independent of both no. of nodes $V$ and no. of features $F$, makes \namemodelnew both node-inductive and feature-inductive. $\hfill\square$.

\textbf{Remark on condition-2:} Node attributes are usually composed of bag-of-words in citation graphs, product categories in e-commerce graphs, and medical diagnoses in case of health-care-related graphs. In such cases, a categorical transformation function $\Phi$ can be learned in the pre-processing stage. For eg., word-embedding methods \cite{fasttext} for the bag of words features, language models\cite{bert} on the textual description of items categories or healthcare-related categories. There also exist specialized product category encoding methods \cite{product_embeddings,metaprod2vec} which utilize skip-gram model on product sequences generated from user sessions. 


\subsection{Expressive Power of \namemodelnew -- Proof of Thm.~\ref{thm:expressivity}}  
\label{app:expressivity}
Similar to other works on analyzing the expressive power of \gnns, we assume that the input feature space $\mathcal{X}$ is countable. Since proposed graph transformation only impacts the features and their information flow during message-passing between graph nodes $\CV$ in $\CG^{alt}$, it is sufficient to show that \namemodelnew produces distinct feature representation for nodes having different features in Phase-1, i.e., after Phase-1 $\h_v^{\ell}\left(v_3\right) \neq \h_v^{\ell}\left(v_4\right) \;\forall\; \ell \in [1\ldots L], v_3,v_4 \in \CV, \;if \exists\; \x_{v_3} \neq \x_{v_4}$. If this holds true, and since Phase-2 \gnn utilizes $\Psi_{\CG}$ itself, Thm.~\ref{thm:expressivity} is proved.
We now need to prove the following lemma:
\begin{lem}[]
\namemodelnew $\Psi_{alt}$ produces distinct transformation for graph nodes $\forall v \in \CV$ in phase -1 given that their input feature vectors are different, i.e., 
\begin{equation}
    \h_v^{\ell}\left(v_3\right) \neq \h_v^{\ell}\left(v_4\right) \\ \;\forall\; \ell \in [1..L], v_3,v_4 \in \CV \; if \;\exists\; \x_{v_3} \neq \x_{v_4}
    \label{eq:distinct_node_features}
\end{equation}
This is true only if following conditions hold:
\begin{enumerate}
    \item $\text{\textsc{Msg}}_{feat}^{\ell}$, $\text{\textsc{Combine}}_{feat}^{\ell}$, $\text{\textsc{Msg}}_{\CG{'}}^{\ell}$ and $\text{\textsc{Combine}}_{\CG'}^{\ell}$ $\forall \ell \in [1\ldots L]$ in equations \ref{eq:fognn_feat_msg}, \ref{eq:fognn_feat_combine}, \ref{eq:fognn_gnode_feat_msg} and \ref{eq:fognn_gnode_feat_combine} are universal function approximators such as MLPs\cite{hornik1989multilayer}.
    \item $\text{\textsc{Aggregate}}_{feat}^{\ell}$ , $\text{\textsc{Aggregate}}_{\CG'}^{\ell}$ should be injective aggregators over \textsc{multisets} i.e. produce different representations of different \textsc{multisets}.
\end{enumerate}
\end{lem}
 If conditions 1 and 2 hold true and since each feature node is initialized with injective transformation in Eq.~\ref{eq:embedding}, for $l=1$ we can see that $\h_v^1\left(v_3\right) \neq \h_v^1\left(v_4\right) \;\forall v_3,v_4 \in \CV \; if \;\exists\; \x_{v_3} \neq \x_{v_4}$ after Phase-1 of \namemodelnew. For $l>1$, after Phase-3, we see that $\h_v^{\ell} \; \forall v \in \CV^{feat}$ will be distinct for all feature nodes due to assumptions 1 and 2. This follows to $\h_v^{\ell}\left(v_3\right) \neq \h_v^{\ell}\left(v_4\right) \; \forall v_3,v_4 \in \CV \; if \;\exists \;\x_{v_3} \neq \x_{v_4} \;\forall \ell > 1$ after Phase-1. This concludes our analysis. $\hfill\square$

\subsection{Proof of Thm~\ref{thm:continuousexpressivity}}
\label{app:continuousexpressivity}

\begin{proof}
{As per Eq.~\ref{eq:fognn_feat_msg}, let $\h_i^{0}\in\mathbb{R}^d$ the representation of feature node corresponding to dimension $i$, which is also learnable, be a one-hot encoding where dimension $i$ is $1$, and rest are $0$. $\h_v^{0}$ is a zero-vector inconsequential to following analysis, and  $e_{iv}$ is the value corresponding to $i^{th}$ dimension of $\x_v$. With these inputs, let us assume $\text{\textsc{Msg}}^{\ell}_{feat}$ is such that it computes messages of the form $\m_v^{1}(i)\in \mathbb{R}^{2d}$ where the first $d$ dimensions are $\h_i^{0}$, the rest of the dimensions have value $\x_v[i]$, i.e., 
$\forall k:d+1\leq k\leq 2d, \m_v^1(i)[k]=x_v[i]$. The learning task is, therefore, to learn the $\text{\textsc{AGGREGATE}}^{\ell}_{feat}$ function $f(\{\!\!\{\m_v^1(i)\}\!\!\})=\x_v$, i.e., recover the original feature vector from the messages received from feature nodes in the allotropic graph. Examining the messages $\m_v^1(i)$ it is clear that, $\x_v[i]=\m_v^1(i)[i] \times \m_v^1(i)[d+i]$. From the universal approximation theorem, an MLP can learn this function; hence, \namemodelnew can recover the original feature space from the allotropic representation.}
\end{proof}

 
\subsection{Complexity Analysis}
\label{app:complexity}
To reduce this computational burden, we bound the number of features by $S^{\CV^{feat}}$ during message aggregation in Phase-1 (Eqs.~\ref{eq:fognn_feat_msg} and \ref{eq:fognn_feat_msg_agg}). Similarly in Phase-3, we bound the number of graph nodes to compute the feature node embedding by $S^{\CV}$. This leads to $O\left(\prod_{i=1}^L \left( S_i\times{S^{\CV^{feat}}} + S^{\CV} \times S^{\CV^{feat}} \right)\right)$ bound on time and space complexities for generating embedding for every node. $S^{\CV}$ is the no. of graph nodes for computing feature node embeddings and $S^{\CV^{feat}}$ is the no. of feature nodes for computing graph node embeddings. In the datasets we have considered for evaluation, all nodes have large dimensional features, but they are highly sparse. For eg., in \textit{Cora} out of $1433$ features, on average $18$ features have value $1$ for all nodes. Similarly, all feature nodes on average have a value of $1$ in $34$ graph nodes. Thus, connecting graph nodes with only those feature nodes having value $1$ and vice-versa results in a low-computational overhead. We perform sampling in case of large scale datasets eg. \textit{Physics}, where each graph node has on average $34$ features having value $1$ 
out of $8415$ features and each feature node has on average $135$ graph nodes.

\subsection{Extension to Continual Learning for Dynamic Graphs}
\label{app:continual}
To perform elastic weight consolidation, we randomly sample a small set of training nodes ${U} \in \CV_{t_l}$ from the graph. Then, we compute the importance of model weights on the loss of ${U} -\Delta \CV_{t_l}$ where $\Delta \CV_{t_l}$ is the set of affected training nodes at time $t$. Specifically, 
\vspace{-0.20in}
\begin{center}
$
\Omega_{w}=\mathbb{E}_{\left(v\right) \sim \left({U} -\Delta \CV_{t_l}\right)}\left[\left(\frac{\delta \mathcal{L}\left(v\right)}{\delta \Theta_{w}}\right)^{2}\right]
 $       
\end{center}

Here, $\Theta$ refers to the model parameters of \namemodelnew, $\Omega_w$ refers to importance of the $w^{th}$ weight parameter. The  term $\frac{\delta \mathcal{L}\left(v\right)}{\delta \Theta_{w}}$ calculates the  gradient of the loss on unaffected nodes with respect to the  parameter $w$. When the parameter update is to take place with respect to the updated data $\Delta \CG_{t}^{\alt}$, we  penalize updates to the weights that are important for the representative sample of nodes that were not updated in the latest timestamp $t$ using $\Omega_w$ calculated above. We accomplish it by the below loss function. 
\vspace{-0.05in}
\begin{equation}
\label{eq:continual_update}
\mathcal{L}_{cont} = \sum_{v \in \Delta \CV_{t_l}}\mathcal{L}\left(v\right) + \sum_{w} \frac{\lambda}{2} \Omega_{w}\left(\Theta^{w}_{t}-\Theta^{w}_{t-1}\right)^{2}
\end{equation}
The first term refers to the loss computed on the updated set of training nodes in the current timestamp. The second term is a quadratic penalty term on the difference between the parameters for the new timestamp and the previous timestamp. We also observe that we only need to store the current model parameters and model parameters of previous phase $t-1$, as evident from Eq.~\ref{eq:continual_update}. $\lambda$ is a hyper-parameter reflecting how important the unaffected portion of the current graph $\CG_t$ is compared to the updated portion of the graph.

\vspace{-0.10in}
\subsection{Experimental Environment}
\label{app:environment}
\vspace{-0.05in}
All experiments are performed on an Intel Xeon Gold 6248 processor with 80 cores, 1 Tesla V-100 GPU card with 32GB GPU memory, and 377 GB RAM with Ubuntu 18.04.  We perform a $60\%{-}20\%{-}20\%$ data split for train-test-validation. These splits are generated at random. In all experiments, we have used 2 layers of message-passing and trained \namemodelnew using the Adam optimizer with a learning rate of $0.0001$ and choose the model based on the best validation loss. All experiments have been executed $5$ times. We report the mean and standard deviations. Standard deviation below $0.01$ have been approximated to $0$ in node classification results.

\subsection{Datasets}
\label{app:datasets}
Cora~\cite{sslge}, CiteSeer~\cite{sslge} and DBLP~\cite{tang2008arnetminer} are citation graphs where each node is a paper and the edge manifests a citation. The node   labels represent the research category.  In Cora and CiteSeer, the node attributes contain a bag of words of the paper text and in DBLP, the node attributes contain bag of words of keywords~\cite{tang2008arnetminer}.   We also use a large scale graph Physics~\cite{shchur2018pitfalls} to show the scalability capabilities of \namemodelnew. Physics is a co-authorship graph where each node is an author and edges represent if two nodes co-authored a paper. Node attributes are bag-of-words of authors' papers. The task is to map each author to its corresponding research area. These graphs are homophilic. We also use a heterophilic dataset, Actor\cite{geomgcn} to evaluate \namemodelnew. The actor is a co-occurrence graph of actor nodes on the same Wikipedia page. Node features are bags of words from Wikipedia pages, and node labels are actor categories from Wikipedia pages. 

\subsection{Impact of 3-phase Message Passing Compared to Vanilla Message Passing on Transformed Graph}
\label{app:3phase}
\begin{table}[h!]
\caption{Performance of \namemodelnew in node classification task when message passing is performed in standard mode on $\CG^{alt}$ on full dataset}
\centering
\resizebox{4in}{!}{%
\begin{tabular}{p{2.2cm}ccc}
\toprule
\textbf{Method} & \textbf{Cora} & \textbf{CiteSeer} & \textbf{Actor}   \\
\midrule
{Traditional} &{$81.36\pm0.80$} & {$66.72\pm1.40$} & {$36.20 \pm 0.29$} \\ 
 \namemodelnew & {$\mathbf{87.6\pm0.73}$} & {$\mathbf{73.90\pm0.84}$} & {$\mathbf{38.90 \pm 0.84}$} \\ 
\bottomrule
\end{tabular}
}
\label{tab:hetero_grafenne}
\end{table}

Table~\ref{tab:hetero_grafenne} shows the importance of the proposed 3-phased message passing framework. Specifically, we use \graphsage on the allotropic graph instead of the proposed 3-phased message passing. As visible, there is a significant drop in quality.

\clearpage
\subsection{Additional Results}
\begin{table}[h!]
\vspace{-0.10in}
\caption{AUCROC  of GRAFENNE (GAT) and GRAFENNE
(GIN) with benchmark GNNs, GAT and GIN on link-prediction task}
\vspace{-0.15in}
\centering
\resizebox{0.8\textwidth}{!}{%
\begin{tabular}{llccc}

\toprule
\textbf{Dataset} & \textbf{Method} & \boldmath{$p=0$} & \boldmath{$p=0.5$} & \boldmath{$p=0.99$}\\ 
\midrule

\multirow{6}{*}{Cora} & \graphsage & {$0.86 \pm0.002$} & $0.84 \pm0.002$ &  $0.7523 \pm0.05$\\

 & \namemodelnew & \boldmath{$0.8780 \pm0.004$} & \boldmath{$0.8501 \pm0.005$}   & \boldmath{$0.8015 \pm0.006$}\\
\cmidrule{2-5}

  & \gat & $0.8681 \pm0.0027$ & $0.8316 \pm0.0010$  & $0.7468 \pm0.0024$\\

 & \namemodelnew(\gat) & \boldmath{$0.8765 \pm0.002$} & \boldmath{$0.8392 \pm0.0047$}  & \boldmath{$0.7841 \pm0.005$}\\
\cmidrule{2-5}

  & \gin & {$0.8552 \pm0.0022$} & $0.8267 \pm0.0037$  & $0.7399 \pm0.0051$\\

 & \namemodelnew(\gin) & \boldmath{$0.8591 \pm0.007$} & \boldmath{$0.8301 \pm0.008$}   & \boldmath{$0.7692 \pm0.011$}\\

\cmidrule{1-5}
\multirow{6}{*}{CiteSeer} & \graphsage & $0.8251 \pm0.005$ & $0.7617 \pm0.001$  & $0.7223 \pm0.002$\\
 & \namemodelnew & \boldmath{$0.8681 \pm0.010$} & \boldmath{$0.8047 \pm0.010$}  & \boldmath{$0.7378 \pm0.008$}\\
\cmidrule{2-5}

 & \gat & $0.8123 \pm0.0011$ & $0.7789 \pm0.0008$ & $0.7205 \pm0.005$\\
 & \namemodelnew(\gat) & \boldmath{$0.8546 \pm0.008$} & \boldmath{$0.8042 \pm0.005$} &  \boldmath{$0.7268 \pm0.010$}\\
\cmidrule{2-5}

 & \gin & $0.8319 \pm0.0031$ & $0.7767 \pm0.0018$  & $0.7203 \pm0.007$\\
 & \namemodelnew(\gin) & \boldmath{$0.8720 \pm0.0058$} & \boldmath{$0.8046 \pm0.0040$} &  \boldmath{$0.7228 \pm0.0003$}\\

\cmidrule{1-5}

\multirow{4}{*}{Actor} & \graphsage & $.6569 \pm0.0$ & $0.7029 \pm0.0026$  & {$0.6969 \pm0.003$}\\
 & \namemodelnew & \boldmath{$0.7047 \pm0.005$} & \boldmath{$0.7021 \pm0.0034$} &  \boldmath{$0.7029 \pm0.0026$}\\
\cmidrule{2-5}

 & \gat & \boldmath{$.7436 \pm0.0016$} & $0.7436 \pm0.0025$ &  {$0.6721 \pm0.0035$}\\
 & \namemodelnew(\gat) & {$0.7142 \pm0.006$} & \boldmath{$0.7090 \pm0.006$} &  \boldmath{$0.6942 \pm0.004$}\\
\cmidrule{2-5}

& \gin & $0.7549 \pm0.0$ & \boldmath{$0.7995 \pm0.0$}  & \boldmath{$0.7856 \pm0.0$}\\
 & \namemodelnew(\gin) & \boldmath{$0.7982 \pm0.0$} & $0.7905 \pm0.0013$ &  $0.7539 \pm0.0$\\
\bottomrule
\end{tabular}%
}
\vspace{-0.20in}
\label{tab:res:link_pred} 
\end{table}

\subsection{Impact of Feature Translation}

{We translate features in Cora dataset by a factor of $10$ in the node classification task. In Table~\ref{app:tab:translate} we observe that the performance of \namemodelnew remain intact. $p$ represents the ratio of features deleted per node as defined in Sec~\ref{exp:emipirical_eval}(Empirical evaluation.).}


\begin{table}[h]
\centering
\begin{tabular}{lllll}
 \toprule
                    &  & $p=0$         & $p=0.5$        & $p=0.99$       \\ \midrule
\namemodelnew            &  & $87.6\pm0.73$ & $84.35\pm0.27$ & $78.85\pm0.29$ \\
\namemodelnew(Translate) &  & $87.7\pm0.81$ & $84.03\pm0.31$ & $78.78\pm0.99$ \\
\bottomrule
\end{tabular}
\caption{{Impact of feature translation (by a factor of 10) on the Cora dataset.}}
\label{app:tab:translate}
\end{table}

\subsection{Comparison with \fate ~\cite{wu2021towards}  }
\label{app:comp_FATE}



As we  explain below, \fate tackles a different problem, is significantly different in methodology and consequently, when adapted for our problem, generates substantially inferior results.

\textbf{Difference in problem formulation:} The input to our problem is a graph where nodes are annotated with feature vectors. In \fate, the input does not include a graph. \fate takes as input just a set of feature vectors. In addition, \fate also assumes the feature vectors to be a set of attributes (represented as one-hot encoding). In our problems, the feature vectors may represent either attributes or continuous-valued.

\textbf{Difference in methodology:} For feature adaptation, \fate forms a bipartite graph where the two sets of nodes correspond to original data points and feature values. While we also form a data(node)-feature graph, the methodology is dramatically different.

    \begin{enumerate}
        \item {\textbf{Structure of graph representation:} \namemodelnew is not a bipartite graph as there are node-node edges in addition to node-feature edges (Recall Fig. 1).}

        \item {\textbf{Modeling feature values:} Since \fate assumes attributed feature vectors, which are one-hot encodings, an edge exists in the bipartite graph from the original data point to a feature-node if the feature (attribute) is present in that data point. This design is not adequate for our problem since features could be continuous-valued. Hence, in our allotropic graph construction, the edges from feature nodes to data nodes are weighted indicating the feature value in the node.}

        \item {\textbf{Handling continuous-value data:} \fate does discuss strategies to adapt their methodology to continuous-valued features by discretizing the feature space into bins. This solution is not adequate since:}

        \begin{enumerate}
        \item  {It’s not clear what should be the bin-width.}

        \item {More importantly, binning feature spaces and treating them as discrete values in the form of one-hot encodings distort the notion of similarity in the original feature space. Specifically, let’s assume a bin width of 25 on a feature ranging from 0 to 100. A value in bins 0-25 is more similar to a value in 26-50 than to one in 76-100. This semantics gets lost when discretized since \fate treats every two bin values as either being the same or different (as in items drawn from a set).}

        \item {Finally, \fate results in $f\times m$ feature-nodes where $f$ is the number of features and $m$ is the number of feature values (bins) per feature on average. In contrast, the proposed work generates $f$ feature nodes. Thus, \fate ~\cite{wu2021towards} leads to higher storage and computation overheads.}
        \end{enumerate}

        \item {\textbf{Message-passing scheme:} Since the input data in \fate is not a graph and the input is assumed to be attributed feature vectors, the message-passing scheme’s primary objective is to learn co-occurrence correlation across features. The task in our case is significantly more complex. Specifically, we (1) need to learn feature co-occurrence patterns, (2) continuous-valued feature imputation as a function of topology, and (3) the objective function (such as node classification, link prediction, etc) as a joint function of topology and features. Owing to the difference in objectives, while \fate decouples the objective task (Ex. classification) from the feature adaptation task. In contrast, we learn feature adaptation and the end objective in an end-to-end manner. Furthermore, while \fate directly uses message passing mechanism of GNNs on the bipartite graph, we have devised a 3-phased message passing framework on the transformed graph due to the more complex modeling needs. As evident from Table F in Appendix Sec. G, the 3-phased message passing obtains superior performance.}
    \end{enumerate}

{\textbf{Empirical evaluation:} We have added \fate~\cite{wu2021towards} as a baseline in the table below. $p$ represents the ratio of features deleted per node as defined in Sec~\ref{exp:emipirical_eval}(Empirical evaluation.). The results are presented for node classification. To adapt \fate for our task where the input is a graph, in addition to the bipartite graph, we added node-node edges as in our construction. Yet, \fate produces significantly inferior results, due to the issues outlined above.}

\begin{table}[h]
\scalebox{0.82}{
\begin{tabular}{lllllclllll}
\toprule
\textbf{Dataset\textbackslash{}Method} &                            & \multicolumn{1}{c}{\namemodelnew} &                               &  & \multicolumn{1}{l}{} &  &                            & \fate                     &                               &  \\ \midrule
                                       & \multicolumn{1}{c}{$p=0$} & \multicolumn{1}{c}{$p=0.5$}  & \multicolumn{1}{c}{$p=0.99$} &  &                      &  & \multicolumn{1}{c}{$p=0$} & \multicolumn{1}{c}{$p=0.5$} & \multicolumn{1}{c}{$p=0.99$} &  \\ \midrule
Cora                                   & $87.6\pm0.73$              & $84.35\pm0.27$                & $78.85\pm0.29$                &  & |                     &  & $71.11\pm1.11$             & $52.43\pm1.21$               & $30.42\pm0.22$                &  \\ \midrule
Actor                                  & $38.9\pm0.84$              & $35.02\pm0.21$                & $23.97\pm0.58$                &  & |                     &  & $34.42\pm0.11$             & $32.25\pm0.93$               & $22.57\pm0.15$                &  \\ \midrule
CiteSeer                               & $73.9\pm0.84$              & $72.91\pm0.95$                & $64.29\pm1.02$                &  & |                     &  & $69.7\pm0.78$              & $61.11\pm0.06$               & $22.69\pm0.51$                &  \\ \midrule
Physics                                & $97.02\pm0.05$             & $96.23\pm0.23$                & $94.49\pm0.18$                &  & |                     &  & $96.34\pm0.09$             & $92.87\pm0.22$               & $53.55\pm0.15$                &  \\ \bottomrule
\end{tabular}
}
\caption{{Comparison of \namemodelnew against \fate at different missing rates on the node classification task.}}
\end{table}

\end{document}